\documentclass[11pt]{article}
\usepackage[margin=1in]{geometry}
\usepackage[T1]{fontenc}
\usepackage{lmodern}
\usepackage{microtype}
\usepackage{amsmath,amssymb}
\usepackage{booktabs}
\usepackage{graphicx}
\usepackage{multirow}
\usepackage{xcolor}
\usepackage{tikz}
\usetikzlibrary{arrows.meta,positioning,calc,fit,backgrounds}
\usepackage{pgfplots}
\pgfplotsset{compat=1.18}
\usepgfplotslibrary{groupplots}
\usepackage[numbers,sort&compress]{natbib}
\usepackage{caption}
\usepackage[hidelinks]{hyperref}
\usepackage{enumitem}
\setlist{nosep,leftmargin=*}

\newcommand{\wer}[2]{#1\,/\,#2}

\title{Gabor Gabor: E-Gaborformer\\[4pt]\large A Strictly Causal Speech Recognizer with Gabor-Dictionary Convolutions and Delta-Rule Memory}
\author{%
\begin{tabular}{c@{\hspace{2.2em}}c@{\hspace{2.2em}}c@{\hspace{2.2em}}c}
Nir Krakowski & Tamir Cohen & Ohad Zohar & Chen Zarfati \\
\small\texttt{nir@deepdub.ai} & \small\texttt{tamir.cohen@deepdub.ai} & \small\texttt{ohad@deepdub.ai} & \small\texttt{chen@deepdub.ai}
\end{tabular}\\[6pt]
Deepdub\\
\small\url{https://deepdub.ai}}
\date{}

\begin{document}
\maketitle

\begin{abstract}
Real-time voice agents need a speech recognizer that never waits for future audio and whose memory per call does not
grow with the length of the call. We present E-Gaborformer, a strictly causal encoder that combines Gated DeltaNet-2
recurrent layers with sliding-window attention in a 3:1 ratio, and builds every temporal convolution from a fixed
dictionary of causal Gabor atoms. Its streaming state is constant, about 19\,MB per call at production scale including the prediction network, where a
causal-attention encoder of the same size accumulates over 1\,GB after ten minutes. In a controlled causal study on
LibriSpeech, E-Gaborformer leaves the initial plateau earlier, reaches a given accuracy up to 18\% sooner, and matches or beats
an E-Branchformer trained with the identical front end and recipe. At production scale, a 0.6B-parameter
Hebrew and English streaming transducer built on E-Gaborformer has 28 to 31\% lower word error rate (WER) than
NVIDIA's Nemotron 3.5 ASR Streaming, a model of the same size, on English voice-agent traffic, and up to 20\% lower Hebrew
WER than our previous production system. The same weights serve strictly causal and 480\,ms-lookahead decoding with no
loss in accuracy, and a two-class formulation of the RNN-T lattice trains the transducer 2.6$\times$ faster with exact
loss and gradients.
\end{abstract}

\section{Introduction}

A voice agent must transcribe the user while they are still speaking, for thousands of concurrent calls. Two properties of the acoustic encoder
matter for this setting: \emph{causality}, so that no output waits for future audio, and a \emph{bounded per-stream
state}, so that serving cost does not grow with call length.

Most strong ASR encoders, such as the Conformer~\citep{gulati2020conformer} and the
E-Branchformer~\citep{kim2023ebranchformer}, are bidirectional. They can be made causal by masking future frames, but
causal self-attention still caches every past key and value: an 18-layer, $d{=}1024$ encoder at 80\,ms frames
accumulates about 110\,MB of fp32 cache per minute of audio per caller. Chunked and memory-augmented transformers bound
this cost with explicit truncation~\citep{shi2021emformer,chen2021streaming,noroozi2024stateful}. Linear-attention
recurrences take another route: they compress the past into a fixed-size state that is updated once per
frame~\citep{katharopoulos2020transformers,schlag2021linear}. The delta-rule family, DeltaNet~\citep{yang2024deltanet},
Gated DeltaNet~\citep{yang2025gdn} and Gated DeltaNet-2 (GDN-2)~\citep{hatamizadeh2026gdn2}, has recently closed much of
the gap to attention in language modeling, and hybrid stacks that mix recurrent and attention layers often do better than
either alone~\citep{de2024griffin,lieber2024jamba,ren2024samba}.

We show that a mostly recurrent causal encoder matches attention in speech recognition, and propose
E-Gaborformer. It keeps the E-Branchformer block and changes three things inside it: a 3:1 GDN-2\,/\,sliding-window-attention global
stack, depthwise convolutions built from a fixed Gabor dictionary, and a change to the GDN-2 output gate that keeps the
recurrent branches active from the first update. The result is an encoder with a constant per-call state that trains
faster than its attention-based counterpart and, at production scale, outperforms a state-of-the-art streaming recognizer
of the same size on our English voice-agent traffic.

\paragraph{Contributions.}
\begin{itemize}
  \item \textbf{A hybrid recurrent encoder for causal ASR} (Section~\ref{sec:model}). Three GDN-2 layers for every
  sliding-window attention layer, inside the E-Branchformer block. GDN-2 branches start 50 to 60$\times$ quieter than
  attention branches under CTC; a change to their output gate, at no parameter cost, removes the resulting training
  stalls.
  \item \textbf{Gabor-dictionary depthwise convolutions} (Section~\ref{sec:gabor}). Restricting convolution kernels to a
  structured basis is an established idea in vision~\citep{jacobsen2016structured,qiu2018dcfnet}, and Gabor functions
  have a long history as speech features~\citep{schadler2012spectro,chang2014robust}. We apply a fixed, one-sided
  (causal) Gabor basis to the temporal depthwise convolutions of a speech encoder, trained from the first step. This
  differs from fitting and freezing Gabor functions after training~\citep{orukeet2026}.
  \item \textbf{A controlled causal comparison} (Section~\ref{sec:study}) against an E-Branchformer with an identical
  front end and recipe, with training curves over ten epochs, and a screen of 36 encoder configurations in about 160
  runs that isolates the contribution of each component.
  \item \textbf{A 0.6B-parameter streaming transducer} (Section~\ref{sec:prod}) whose prediction network interleaves GDN-2
  with token attention. On English voice-agent traffic it has 28 to 31\% lower WER than Nemotron 3.5 ASR Streaming, and one
  set of weights serves both strictly causal and 480\,ms-lookahead decoding at equal accuracy.
  \item \textbf{Exact and fast transducer training} (Section~\ref{sec:speed}): a two-class RNN-T lattice with a
  closed-form logit gradient, aligned output layers, tensor-core depthwise convolutions and weight-preserving length
  regrouping, together 2.6$\times$ faster per update with an unchanged objective.
\end{itemize}

\section{Related work}

\paragraph{Encoders for ASR.} The Conformer~\citep{gulati2020conformer} interleaves self-attention with a convolution
module. The Branchformer~\citep{peng2022branchformer} runs attention and a convolutional-gating MLP (cgMLP) in parallel
branches, and the E-Branchformer~\citep{kim2023ebranchformer} adds a depthwise-convolution merge; it matches or exceeds
the Conformer across tasks~\citep{peng2023comparative}. FastConformer~\citep{rekesh2023fastconformer} subsamples
8$\times$ for efficiency.

\paragraph{Streaming recognizers.} Cache-aware streaming~\citep{noroozi2024stateful} trains a FastConformer with limited
right context and reuses cached activations at inference; training with several right-context sizes lets one model serve
several latencies; dual-mode and dynamic-chunk training~\citep{yu2021dualmode,zhang2020unified} pursue the same goal.
Classic streaming recognizers bound the state with LSTM encoders~\citep{he2019streaming} or truncated attention
context~\citep{zhang2020transformer,shi2021emformer}. Nemotron 3.5 ASR Streaming~\citep{nemotron2026asr} applies this recipe to a 0.6B-parameter
multilingual RNN-T with a language prompt. Its attention cache is bounded by a fixed left context of 56 frames per layer.
We use the same multi-lookahead training idea; our contribution is the encoder and prediction network, not the
streaming recipe.

\paragraph{Recurrent and hybrid sequence models.} Linear attention~\citep{katharopoulos2020transformers} and
fast-weight programmers~\citep{schlag2021linear} recast attention as a recurrence with a matrix-valued state. The delta
rule~\citep{yang2024deltanet} replaces additive writes with error-correcting ones, gating adds data-dependent
forgetting~\citep{yang2025gdn}, and GDN-2~\citep{hatamizadeh2026gdn2} decouples erasing from writing with channel-wise
gates. Structured state-space models~\citep{gu2022s4,gu2024mamba}, long convolutions~\citep{poli2023hyena} and hybrids
of recurrent and local-attention layers~\citep{de2024griffin,lieber2024jamba,ren2024samba} show that a few attention
layers recover the exact recall that pure recurrences lack. In speech, Zipformer~\citep{yao2024zipformer} and
state-space or Mamba encoders~\citep{jiang2024speechslytherin} are recent efficient alternatives to the Conformer.

\paragraph{Structured and parametrized filters.} Structured receptive fields~\citep{jacobsen2016structured} and
DCFNet~\citep{qiu2018dcfnet} express CNN kernels as learned combinations of a fixed basis (Gaussian derivatives and
Fourier--Bessel functions). Gabor convolutional networks~\citep{luan2018gabor} modulate learned kernels with Gabor
filters, and GaborNet~\citep{alekseev2019gabornet} learns the parameters of Gabor kernels directly. In speech,
spectro-temporal Gabor filters serve as robust features~\citep{schadler2012spectro} and as fixed CNN
kernels~\citep{chang2014robust}, and SincNet~\citep{ravanelli2018sincnet} and LEAF~\citep{zeghidour2021leaf} learn
band-pass front-end filters through their center frequencies and bandwidths. Orukeet~\citep{orukeet2026} fits a Gabor
function to each nine-tap depthwise kernel of a trained Parakeet model, replaces the best-fitting half by its fit and
freezes it during further training. Our filters are temporal rather than spectro-temporal, act inside every encoder
block rather than at the input, and are constrained to the Gabor span from initialization.

\paragraph{Efficient transducer training.} The RNN-T loss~\citep{graves2012rnnt} requires joint-network outputs at every
(frame, token) node. Function merging~\citep{li2019rnnt} avoids storing full log-softmax outputs, and pruned
RNN-T~\citep{kuang2022pruned} restricts the lattice to a band. Fused cross-entropy kernels avoid materializing large
logit matrices in language models~\citep{wijmans2025cce}.

\section{Preliminaries}
\label{sec:prelim}

\paragraph{The E-Branchformer block.} An E-Branchformer layer~\citep{kim2023ebranchformer} is a macaron block: a
half-step feed-forward layer, a \emph{global} and a \emph{local} branch computed in parallel, a merge, and a second
half-step feed-forward layer. The global branch is multi-head self-attention. The local branch is a cgMLP: a projection
to $2c$ channels, a split into value and gate halves, and a depthwise convolution over time on the gate. The merge
concatenates the two branches, applies a second depthwise convolution and projects back to $d$.

\paragraph{Gated DeltaNet-2.} GDN-2~\citep{hatamizadeh2026gdn2} keeps, per head, a matrix state
$S_t \in \mathbb{R}^{d_k \times d_v}$ ($64\times64$ in all our models) and updates it once per frame:
\begin{equation}
  S_t = \bigl(I - k_t (b_t \odot k_t)^{\!\top}\bigr)\, \mathrm{Diag}\!\left(e^{g_t}\right) S_{t-1}
        + k_t\, (w_t \odot v_t)^{\!\top}, \qquad o_t = S_t^{\top} q_t .
\end{equation}
The decay $g_t \le 0$ forgets channel by channel, the erase gate $b_t$ removes what the current key already stores, and
the write gate $w_t$ selects which value channels are written. Queries, keys and values pass through causal short
convolutions (kernel 4), keys and queries are $\ell_2$-normalized, the decay uses a low-rank projection (rank 64), and
the per-head output passes through a gated root-mean-square normalization (RMSNorm) before the output projection. We use the chunked Triton kernels of
flash-linear-attention~\citep{fla2024,tillet2019triton}.

\paragraph{CTC and RNN-T.} Both training objectives sum the probability of the reference transcript over all of its
alignments to the encoder frames. Connectionist temporal classification (CTC)~\citep{graves2006ctc} adds a blank symbol
and predicts one symbol or blank per frame, independently of the symbols already emitted; we use it in the controlled
study (Section~\ref{sec:study}). In \emph{intermediate CTC}~\citep{lee2021interctc}, a second CTC loss is computed from the
output of an inner encoder layer through the same output layer (self-conditioned CTC~\citep{nozaki2021selfcond} also
feeds that layer's prediction back into the encoder), and the training loss becomes $(1-w)\,L_{\text{CTC}} +
w\,L_{\text{CTC}}^{(\ell)}$; we tried $\ell = 6$ and $w = 0.3$ as an ablation. The recurrent neural network transducer
(RNN-T)~\citep{graves2012rnnt} adds a \emph{prediction network} that reads the symbols emitted so far, and a \emph{joint
network} that combines its output with each encoder frame into a distribution over the vocabulary and blank. Its loss
is the negative log-probability of the transcript summed over the paths through the frame--symbol lattice, in which
blank advances one frame and a symbol advances one position in the transcript. Unlike CTC, the transducer conditions
each output on the previous symbols, and it is the model we deploy (Section~\ref{sec:prod}).

\paragraph{Causality.} We call a model strictly causal if output frame $j$ depends on no input after frame $j$. Output
frame $j$ covers the eight 10\,ms analysis hops ending at $80j$\,ms; with a 25\,ms STFT window it depends on audio up to
$80j + 12.5$\,ms. This requires masked attention, left-padded convolutions, a forward-only recurrence and a subsampling
front end that pads only on the left in time.

\section{E-Gaborformer}
\label{sec:model}

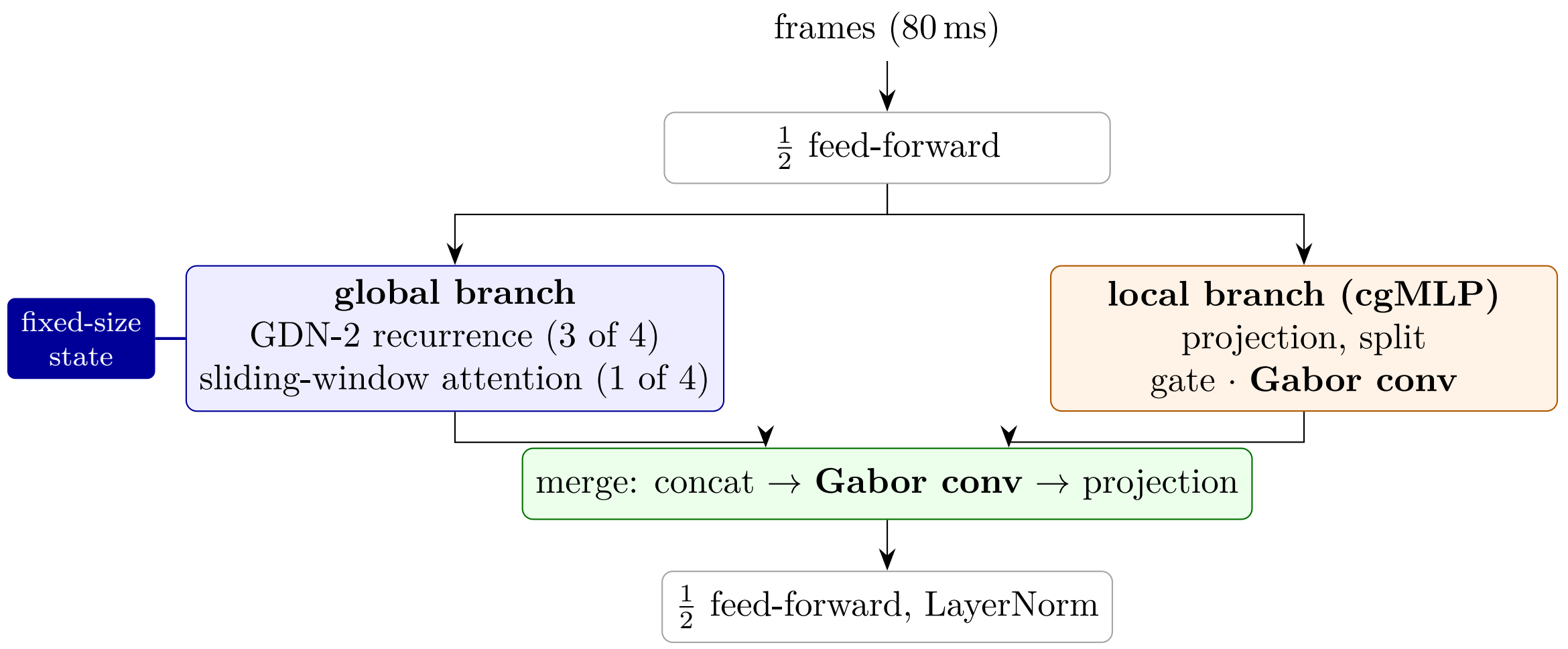
\begin{figure}[t]
\centering
\begin{tikzpicture}[
  font=\small, >={Stealth[length=2.2mm]},
  box/.style={draw=black!35, rounded corners=3pt, minimum height=7mm, align=center, fill=white},
  gl/.style={box, fill=blue!7, draw=blue!60!black},
  lo/.style={box, fill=orange!10, draw=orange!70!black},
  mg/.style={box, fill=green!8, draw=green!45!black}]
\node[box, minimum width=44mm] (ff1) {$\tfrac12$ feed-forward};
\node[gl, below left=8mm and -6mm of ff1, minimum width=50mm] (g) {\textbf{global branch}\\GDN-2 recurrence (3 of 4)\\sliding-window attention (1 of 4)};
\node[lo, below right=8mm and -6mm of ff1, minimum width=50mm] (l) {\textbf{local branch (cgMLP)}\\projection, split\\gate $\cdot$ \textbf{Gabor conv}};
\node[mg, below=26mm of ff1, minimum width=62mm] (m) {merge: concat $\rightarrow$ \textbf{Gabor conv} $\rightarrow$ projection};
\node[box, below=5mm of m, minimum width=44mm] (ff2) {$\tfrac12$ feed-forward, LayerNorm};
\draw[->] (ff1.south) -- ++(0,-3mm) -| (g.north);
\draw[->] (ff1.south) -- ++(0,-3mm) -| (l.north);
\draw[->] (g.south) -- ++(0,-3mm) -| ([xshift=-12mm]m.north);
\draw[->] (l.south) -- ++(0,-3mm) -| ([xshift=12mm]m.north);
\draw[->] (m) -- (ff2);
\draw[<-] (ff1.north) -- ++(0,5mm) node[above]{frames (80\,ms)};
\node[draw=blue!60!black, fill=blue!60!black, text=white, rounded corners=2pt, left=3mm of g, font=\scriptsize, align=center] (s) {fixed-size\\state};
\draw[blue!60!black, thick] (s) -- (g);
\end{tikzpicture}
\caption{One E-Gaborformer block (residual connections omitted). The global branch is a GDN-2 recurrence in three
layers out of four and causal sliding-window attention in the fourth; both depthwise convolutions are Gabor-dictionary
convolutions.}
\label{fig:block}
\end{figure}

E-Gaborformer keeps the E-Branchformer block and changes its global branch, its two depthwise convolutions and the
output gate of the recurrent layers (Figure~\ref{fig:block}).

\subsection{Hybrid global branch}
\label{sec:hybrid}
The global branch is a GDN-2 recurrence in every layer whose index (counted from 1) is not a multiple of four, and
causal sliding-window
attention over the last $W{=}128$ frames (10.24\,s) with rotary position embeddings~\citep{su2024roformer} in every fourth
layer; we write this pattern as GGGS (G: GDN-2, S: sliding-window attention). The recurrence can carry information over
arbitrarily long distances in a fixed-size state, subject to its learned decay; the attention layers provide
exact recall over the recent window, which a compressed state cannot.

\subsection{Gabor-dictionary convolutions}
\label{sec:gabor}
\begin{figure}[t]
\centering
\begin{tikzpicture}
\begin{groupplot}[group style={group size=4 by 4, horizontal sep=4pt, vertical sep=14pt},
 width=0.30\linewidth, height=2.6cm, xmin=-30, xmax=0, ymin=-0.75, ymax=0.75, xtick=\empty, ytick=\empty,
 axis line style={draw=black!25}, title style={font=\scriptsize, yshift=-5pt}]
\nextgroupplot[title={Gaussian $\sigma=1.5$}] \addplot[blue!70!black, thick, mark=none] coordinates {(-30,0.0000) (-29,0.0000) (-28,0.0000) (-27,0.0000) (-26,0.0000) (-25,0.0000) (-24,0.0000) (-23,0.0000) (-22,0.0000) (-21,0.0000) (-20,0.0000) (-19,0.0000) (-18,0.0000) (-17,0.0000) (-16,0.0000) (-15,0.0000) (-14,0.0000) (-13,0.0000) (-12,0.0000) (-11,0.0000) (-10,0.0000) (-9,0.0000) (-8,0.0000) (-7,0.0000) (-6,0.0002) (-5,0.0029) (-4,0.0211) (-3,0.1001) (-2,0.3040) (-1,0.5920) (0,0.7394)};
\nextgroupplot[title={Gaussian $\sigma=4$}] \addplot[blue!70!black, thick, mark=none] coordinates {(-30,0.0000) (-29,0.0000) (-28,0.0000) (-27,0.0000) (-26,0.0000) (-25,0.0000) (-24,0.0000) (-23,0.0000) (-22,0.0000) (-21,0.0000) (-20,0.0000) (-19,0.0000) (-18,0.0000) (-17,0.0001) (-16,0.0002) (-15,0.0004) (-14,0.0011) (-13,0.0025) (-12,0.0055) (-11,0.0113) (-10,0.0218) (-9,0.0396) (-8,0.0673) (-7,0.1075) (-6,0.1614) (-5,0.2276) (-4,0.3016) (-3,0.3753) (-2,0.4388) (-1,0.4819) (0,0.4972)};
\nextgroupplot[title={cos 0.38 Hz}] \addplot[orange!85!black, thick, mark=none] coordinates {(-30,0.0411) (-29,0.0396) (-28,0.0352) (-27,0.0273) (-26,0.0157) (-25,-0.0000) (-24,-0.0195) (-23,-0.0426) (-22,-0.0686) (-21,-0.0964) (-20,-0.1248) (-19,-0.1522) (-18,-0.1769) (-17,-0.1970) (-16,-0.2107) (-15,-0.2164) (-14,-0.2127) (-13,-0.1986) (-12,-0.1736) (-11,-0.1381) (-10,-0.0928) (-9,-0.0393) (-8,0.0204) (-7,0.0837) (-6,0.1475) (-5,0.2086) (-4,0.2639) (-3,0.3105) (-2,0.3457) (-1,0.3677) (0,0.3752)};
\nextgroupplot[title={sin 0.38 Hz}] \addplot[violet, thick, mark=none] coordinates {(-30,0.0309) (-29,0.0437) (-28,0.0574) (-27,0.0714) (-26,0.0849) (-25,0.0968) (-24,0.1060) (-23,0.1114) (-22,0.1118) (-21,0.1062) (-20,0.0938) (-19,0.0741) (-18,0.0470) (-17,0.0128) (-16,-0.0275) (-15,-0.0728) (-14,-0.1210) (-13,-0.1700) (-12,-0.2172) (-11,-0.2600) (-10,-0.2956) (-9,-0.3217) (-8,-0.3361) (-7,-0.3372) (-6,-0.3243) (-5,-0.2971) (-4,-0.2565) (-3,-0.2039) (-2,-0.1416) (-1,-0.0726) (0,0.0000)};
\nextgroupplot[title={cos 0.58 Hz}] \addplot[orange!85!black, thick, mark=none] coordinates {(-30,-0.0070) (-29,-0.0064) (-28,-0.0041) (-27,0.0003) (-26,0.0073) (-25,0.0170) (-24,0.0288) (-23,0.0420) (-22,0.0548) (-21,0.0650) (-20,0.0700) (-19,0.0669) (-18,0.0534) (-17,0.0279) (-16,-0.0099) (-15,-0.0584) (-14,-0.1137) (-13,-0.1705) (-12,-0.2214) (-11,-0.2589) (-10,-0.2755) (-9,-0.2655) (-8,-0.2257) (-7,-0.1568) (-6,-0.0633) (-5,0.0466) (-4,0.1617) (-3,0.2694) (-2,0.3572) (-1,0.4146) (0,0.4346)};
\nextgroupplot[title={sin 0.58 Hz}] \addplot[violet, thick, mark=none] coordinates {(-30,-0.0064) (-29,-0.0107) (-28,-0.0155) (-27,-0.0203) (-26,-0.0243) (-25,-0.0262) (-24,-0.0247) (-23,-0.0183) (-22,-0.0059) (-21,0.0130) (-20,0.0381) (-19,0.0680) (-18,0.0998) (-17,0.1297) (-16,0.1528) (-15,0.1637) (-14,0.1576) (-13,0.1311) (-12,0.0827) (-11,0.0140) (-10,-0.0705) (-9,-0.1632) (-8,-0.2539) (-7,-0.3318) (-6,-0.3859) (-5,-0.4075) (-4,-0.3912) (-3,-0.3361) (-2,-0.2462) (-1,-0.1301) (0,0.0000)};
\nextgroupplot[title={cos 0.89 Hz}] \addplot[orange!85!black, thick, mark=none] coordinates {(-30,0.0000) (-29,0.0001) (-28,0.0002) (-27,0.0003) (-26,0.0003) (-25,0.0002) (-24,-0.0004) (-23,-0.0016) (-22,-0.0036) (-21,-0.0061) (-20,-0.0081) (-19,-0.0081) (-18,-0.0038) (-17,0.0071) (-16,0.0255) (-15,0.0492) (-14,0.0723) (-13,0.0847) (-12,0.0742) (-11,0.0314) (-10,-0.0459) (-9,-0.1474) (-8,-0.2497) (-7,-0.3203) (-6,-0.3274) (-5,-0.2517) (-4,-0.0964) (-3,0.1094) (-2,0.3164) (-1,0.4695) (0,0.5260)};
\nextgroupplot[title={sin 0.89 Hz}] \addplot[violet, thick, mark=none] coordinates {(-30,-0.0000) (-29,-0.0000) (-28,0.0000) (-27,0.0002) (-26,0.0005) (-25,0.0010) (-24,0.0016) (-23,0.0020) (-22,0.0017) (-21,-0.0003) (-20,-0.0046) (-19,-0.0118) (-18,-0.0209) (-17,-0.0294) (-16,-0.0324) (-15,-0.0238) (-14,0.0020) (-13,0.0466) (-12,0.1049) (-11,0.1631) (-10,0.1999) (-9,0.1926) (-8,0.1247) (-7,-0.0044) (-6,-0.1747) (-5,-0.3465) (-4,-0.4706) (-3,-0.5037) (-2,-0.4242) (-1,-0.2425) (0,0.0000)};
\nextgroupplot[title={cos 1.37 Hz}] \addplot[orange!85!black, thick, mark=none] coordinates {(-30,-0.0000) (-29,0.0000) (-28,0.0000) (-27,0.0000) (-26,0.0000) (-25,-0.0000) (-24,-0.0000) (-23,-0.0000) (-22,-0.0000) (-21,-0.0000) (-20,0.0000) (-19,0.0001) (-18,0.0003) (-17,0.0004) (-16,0.0000) (-15,-0.0018) (-14,-0.0056) (-13,-0.0097) (-12,-0.0078) (-11,0.0096) (-10,0.0472) (-9,0.0898) (-8,0.0966) (-7,0.0205) (-6,-0.1460) (-5,-0.3301) (-4,-0.3969) (-3,-0.2406) (-2,0.1104) (-1,0.4747) (0,0.6296)};
\nextgroupplot[title={sin 1.37 Hz}] \addplot[violet, thick, mark=none] coordinates {(-30,-0.0000) (-29,-0.0000) (-28,-0.0000) (-27,0.0000) (-26,0.0000) (-25,0.0000) (-24,0.0000) (-23,0.0000) (-22,-0.0000) (-21,-0.0000) (-20,-0.0000) (-19,-0.0001) (-18,0.0001) (-17,0.0005) (-16,0.0015) (-15,0.0025) (-14,0.0013) (-13,-0.0056) (-12,-0.0204) (-11,-0.0370) (-10,-0.0360) (-9,0.0089) (-8,0.1061) (-7,0.2157) (-6,0.2475) (-5,0.1140) (-4,-0.1814) (-3,-0.4988) (-2,-0.6268) (-1,-0.4361) (0,0.0000)};
\nextgroupplot[title={cos 2.11 Hz}] \addplot[orange!85!black, thick, mark=none] coordinates {(-30,0.0000) (-29,0.0000) (-28,-0.0000) (-27,-0.0000) (-26,-0.0000) (-25,0.0000) (-24,0.0000) (-23,0.0000) (-22,-0.0000) (-21,-0.0000) (-20,-0.0000) (-19,0.0000) (-18,0.0000) (-17,0.0000) (-16,-0.0000) (-15,-0.0000) (-14,-0.0000) (-13,0.0000) (-12,0.0002) (-11,0.0005) (-10,-0.0010) (-9,-0.0073) (-8,-0.0114) (-7,0.0192) (-6,0.0954) (-5,0.0991) (-4,-0.1360) (-3,-0.4446) (-2,-0.3086) (-1,0.3430) (0,0.7426)};
\nextgroupplot[title={sin 2.11 Hz}] \addplot[violet, thick, mark=none] coordinates {(-30,-0.0000) (-29,0.0000) (-28,0.0000) (-27,0.0000) (-26,-0.0000) (-25,-0.0000) (-24,-0.0000) (-23,0.0000) (-22,0.0000) (-21,0.0000) (-20,-0.0000) (-19,-0.0000) (-18,-0.0000) (-17,0.0000) (-16,0.0000) (-15,0.0000) (-14,-0.0000) (-13,-0.0001) (-12,-0.0000) (-11,0.0007) (-10,0.0027) (-9,0.0010) (-8,-0.0185) (-7,-0.0487) (-6,-0.0086) (-5,0.1751) (-4,0.3125) (-3,0.0199) (-2,-0.5928) (-1,-0.7191) (0,0.0000)};
\nextgroupplot[title={cos 3.25 Hz}] \addplot[orange!85!black, thick, mark=none] coordinates {(-30,0.0000) (-29,-0.0000) (-28,-0.0000) (-27,0.0000) (-26,0.0000) (-25,-0.0000) (-24,0.0000) (-23,0.0000) (-22,-0.0000) (-21,-0.0000) (-20,0.0000) (-19,0.0000) (-18,-0.0000) (-17,-0.0000) (-16,0.0000) (-15,0.0000) (-14,-0.0000) (-13,-0.0000) (-12,0.0000) (-11,0.0000) (-10,-0.0000) (-9,-0.0000) (-8,0.0001) (-7,0.0005) (-6,-0.0062) (-5,-0.0089) (-4,0.0963) (-3,0.0467) (-2,-0.4973) (-1,-0.0460) (0,0.8597)};
\nextgroupplot[title={sin 3.25 Hz}] \addplot[violet, thick, mark=none] coordinates {(-30,0.0000) (-29,0.0000) (-28,-0.0000) (-27,-0.0000) (-26,0.0000) (-25,-0.0000) (-24,-0.0000) (-23,0.0000) (-22,0.0000) (-21,-0.0000) (-20,-0.0000) (-19,0.0000) (-18,0.0000) (-17,-0.0000) (-16,-0.0000) (-15,0.0000) (-14,0.0000) (-13,-0.0000) (-12,-0.0000) (-11,0.0000) (-10,0.0000) (-9,-0.0000) (-8,-0.0001) (-7,0.0013) (-6,0.0030) (-5,-0.0354) (-4,-0.0303) (-3,0.3160) (-2,0.0772) (-1,-0.9444) (0,0.0000)};
\nextgroupplot[title={cos 5.00 Hz}] \addplot[orange!85!black, thick, mark=none] coordinates {(-30,0.0000) (-29,-0.0000) (-28,0.0000) (-27,0.0000) (-26,-0.0000) (-25,0.0000) (-24,-0.0000) (-23,0.0000) (-22,0.0000) (-21,-0.0000) (-20,0.0000) (-19,-0.0000) (-18,0.0000) (-17,0.0000) (-16,-0.0000) (-15,0.0000) (-14,-0.0000) (-13,0.0000) (-12,0.0000) (-11,-0.0000) (-10,0.0000) (-9,-0.0000) (-8,0.0000) (-7,0.0000) (-6,-0.0000) (-5,0.0003) (-4,-0.0042) (-3,0.0149) (-2,0.0739) (-1,-0.5051) (0,0.8598)};
\nextgroupplot[title={sin 5.00 Hz}] \addplot[violet, thick, mark=none] coordinates {(-30,0.0000) (-29,0.0000) (-28,-0.0000) (-27,0.0000) (-26,-0.0000) (-25,0.0000) (-24,0.0000) (-23,-0.0000) (-22,0.0000) (-21,-0.0000) (-20,0.0000) (-19,0.0000) (-18,-0.0000) (-17,0.0000) (-16,-0.0000) (-15,0.0000) (-14,0.0000) (-13,-0.0000) (-12,0.0000) (-11,-0.0000) (-10,0.0000) (-9,0.0000) (-8,-0.0000) (-7,0.0000) (-6,-0.0000) (-5,0.0000) (-4,0.0070) (-3,-0.1057) (-2,0.5237) (-1,-0.8453) (0,0.0000)};
\end{groupplot}
\end{tikzpicture}
\caption{The 16 fixed causal Gabor atoms (kernel 31, lag $-30 \ldots 0$ left to right; frequencies at 80\,ms frames).
Each channel's kernel is a learned linear combination of them.}
\label{fig:atoms}
\end{figure}
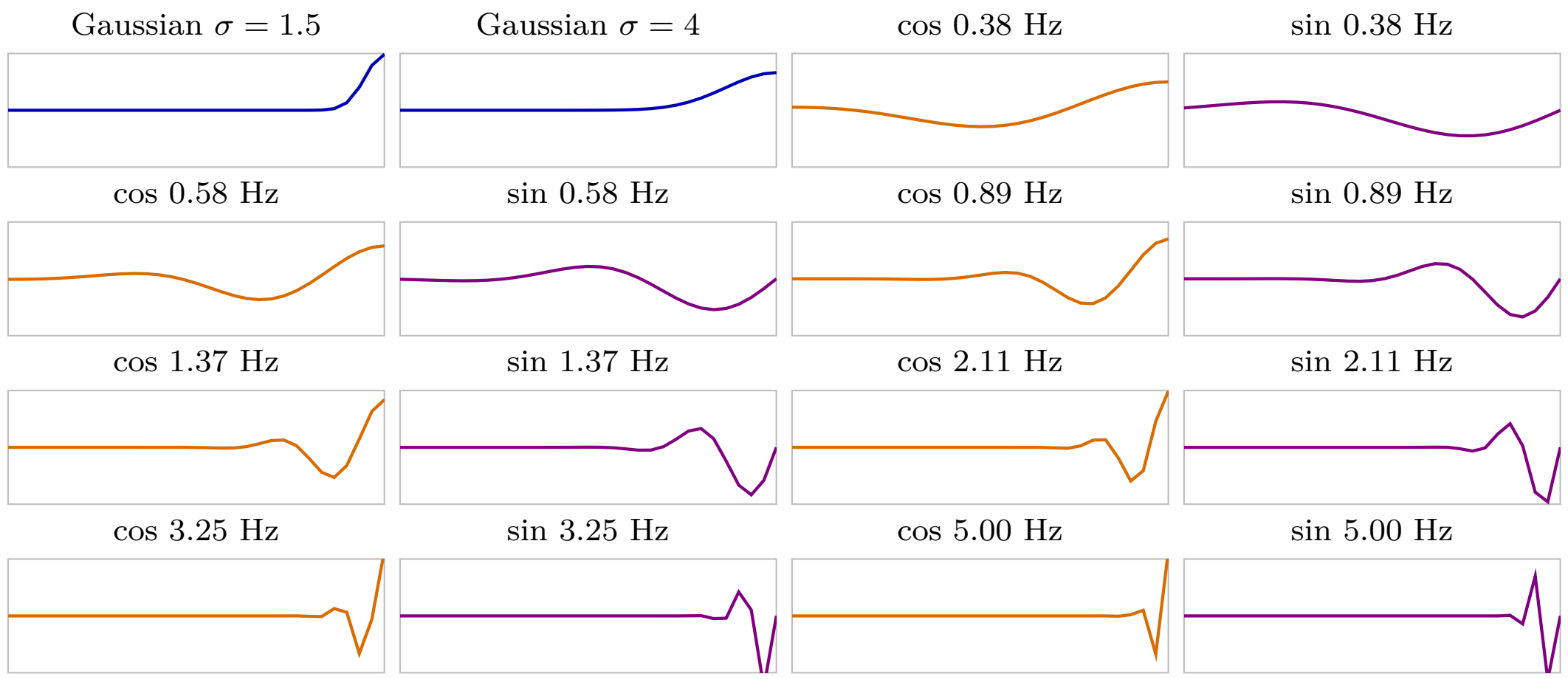

A depthwise convolution with kernel size $K$ applies, per channel $c$, a kernel $h_c \in \mathbb{R}^K$. Following Gabor's time--frequency atoms~\citep{gabor1946theory}, we
restrict $h_c$ to the span of a fixed dictionary $\{\phi_m\}_{m=1}^{M}$:
\begin{equation}
  h_c[\tau] = \sum_{m=1}^{M} a_{c,m}\, \phi_m[\tau], \qquad \tau = -(K-1), \dots, 0,
\end{equation}
and learn only the coefficients $a_{c,m}$ and a bias. The dictionary has $M{=}16$ unit-norm atoms over lags $\tau \le 0$
(the past): two Gaussians $\exp(-\tau^2/2\sigma^2)$ with $\sigma \in \{1.5, 4\}$ frames, and seven wave packets at
frequencies $f_k$ log-spaced from 0.03 to 0.4 cycles per frame, each as a cosine and a sine,
\begin{equation}
  \phi^{\cos}_k[\tau] \propto e^{-\tau^2 / 2\sigma_k^2} \cos(2\pi f_k \tau), \quad
  \phi^{\sin}_k[\tau] \propto e^{-\tau^2 / 2\sigma_k^2} \sin(2\pi f_k \tau), \quad
  \sigma_k = \mathrm{clip}(0.5/f_k,\, 1,\, (K-1)/2),
\end{equation}
so that any phase is reachable by linear mixing (Figure~\ref{fig:atoms}). The envelope peaks at the current frame and
decays into the past, so every kernel is causal by construction. With $K{=}31$ at 80\,ms frames, a filter spans
2.48\,s, and the wave packets are centered between 0.375 and 5\,Hz, the range of phrasal and slower syllabic
modulation~\citep{ding2017temporal}. In the controlled study, whose Gabor ablations ran at 40\,ms frames, the same atoms
cover 0.75 to 10\,Hz.

Unlike filters whose center frequency enters nonlinearly~\citep{ravanelli2018sincnet,zeghidour2021leaf}, the kernel is
linear in its coefficients. It uses 16 parameters per channel instead of $K$ and is built in fp32 before being cast to the
activation type. The constraint acts as a soft smoothness prior: the dictionary concentrates energy below about 0.4
cycles per frame. It is not a hard limit, because the one-sided truncation leaks energy from the highest-frequency atoms
above that band. The raw dictionary is ill-conditioned (Gram condition number about $8 \times 10^6$, effectively about 12
independent directions); all models use it as is, and the coefficients are trained with AdamW. The gate convolution's
coefficients start near zero with a bias of one, so that the gate starts open; the merge convolution's start random.

\subsection{Training stalls of GDN-2 branches}
\label{sec:silent}
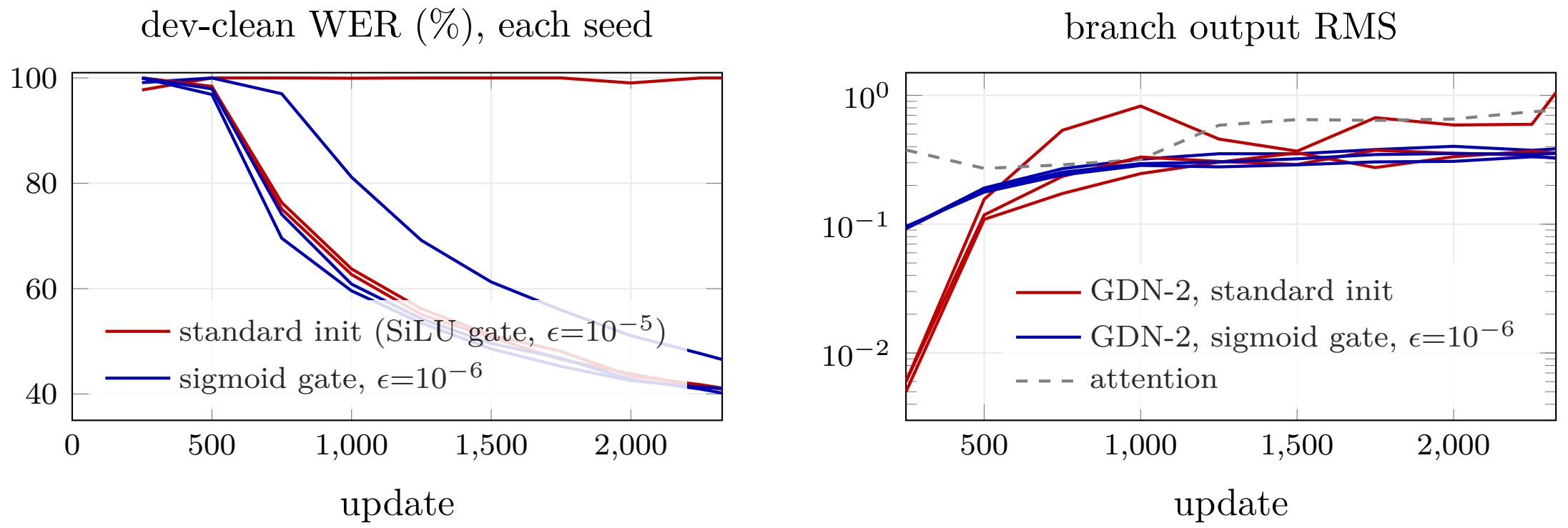
\begin{figure}[t]
\centering
\begin{tikzpicture}
\begin{groupplot}[group style={group size=2 by 1, horizontal sep=17mm},
  width=0.46\linewidth, height=4.8cm, xmin=0, xmax=2327, xtick={0,500,1000,1500,2000},
  xlabel={update}, grid=major, grid style={black!8}, tick label style={font=\scriptsize},
  label style={font=\small}, title style={font=\small, yshift=-1mm}, no markers]
\nextgroupplot[title={dev-clean WER (\%), each seed}, ymin=35, ymax=101,
  legend cell align=left, legend style={draw=none, fill=white, fill opacity=0.85, font=\scriptsize, at={(0.03,0.03)}, anchor=south west}]
\addplot[red!75!black, thick] table[x=step, y=wc] {figs/init_std_2022.dat}; \addlegendentry{standard init (SiLU gate, $\epsilon{=}10^{-5}$)}
\addplot[blue!70!black, thick] table[x=step, y=wc] {figs/init_loud_2022.dat}; \addlegendentry{sigmoid gate, $\epsilon{=}10^{-6}$}
\addplot[red!75!black, thick, forget plot] table[x=step, y=wc] {figs/init_std_2023.dat};
\addplot[red!75!black, thick, forget plot] table[x=step, y=wc] {figs/init_std_2024.dat};
\addplot[blue!70!black, thick, forget plot] table[x=step, y=wc] {figs/init_loud_2023.dat};
\addplot[blue!70!black, thick, forget plot] table[x=step, y=wc] {figs/init_loud_2024.dat};
\nextgroupplot[title={branch output RMS}, ymode=log, ymin=0.003, ymax=1.5, xmin=250,
  legend cell align=left, legend style={draw=none, fill=white, fill opacity=0.85, font=\scriptsize, at={(0.97,0.03)}, anchor=south east}]
\addplot[red!75!black, thick] table[x=step, y=G] {figs/init_std_2022.dat}; \addlegendentry{GDN-2, standard init}
\addplot[blue!70!black, thick] table[x=step, y=G] {figs/init_loud_2022.dat}; \addlegendentry{GDN-2, sigmoid gate, $\epsilon{=}10^{-6}$}
\addplot[black!50, thick, dashed] table[x=step, y=S] {figs/init_std_2022.dat}; \addlegendentry{attention}
\addplot[red!75!black, thick, forget plot] table[x=step, y=G] {figs/init_std_2023.dat};
\addplot[red!75!black, thick, forget plot] table[x=step, y=G] {figs/init_std_2024.dat};
\addplot[blue!70!black, thick, forget plot] table[x=step, y=G] {figs/init_loud_2023.dat};
\addplot[blue!70!black, thick, forget plot] table[x=step, y=G] {figs/init_loud_2024.dat};
\end{groupplot}
\end{tikzpicture}
\caption{Standard (SiLU) and sigmoid GDN-2 output gate in the causal 1:1 hybrid (three seeds each, 40\,ms front end).
Left: one SiLU-gate seed never leaves the blank plateau. Right: output RMS of the GDN-2 and attention branches,
logged at every evaluation.}
\label{fig:init}
\end{figure}

Our first hybrid models used the standard GDN-2 configuration, a SiLU output gate and an RMSNorm epsilon of $10^{-5}$.
Four of the 18 causal runs with GDN-2 layers in round 2 of our screen (Section~\ref{sec:explore}) never left the
initial CTC plateau within an epoch, the phase in which the model emits only blanks and the loss stays
flat~\citep{zeyer2021peaky} (Table~\ref{tab:explore}). At the first evaluation, after 250 updates, the GDN-2 branch output
had a root-mean-square (RMS) value of 0.005 to 0.006, against 0.28 to 0.38 for the attention branch of the same model,
50 to 60$\times$ smaller (Figure~\ref{fig:init}, right). We attribute this to the output path: SiLU of a near-zero gate
pre-activation is close to zero, and the recurrence output is small enough to fall below the normalization epsilon, so
RMSNorm does not rescale it. We hypothesize that the rest of the network then learns to work without the recurrent
branch; in the stalled seed, the branch became louder later without the run leaving the plateau.

We replace the SiLU output gate by a sigmoid, which starts at 0.5, and lower the RMSNorm epsilon to $10^{-6}$. Neither
change adds parameters. The GDN-2 branch then starts at an RMS of about 0.09, and
all 36 later causal runs with GDN-2 layers left the plateau (0 stalls against 4 of 18 before), after 750 or 1{,}000
updates against 750 for the E-Branchformer (Figure~\ref{fig:init}, left). The fix changes the stall rate; seeds that did
not stall reached similar WER with either gate. All results after round 2 use the sigmoid gate.

\subsection{Streaming state size}
\label{sec:state}
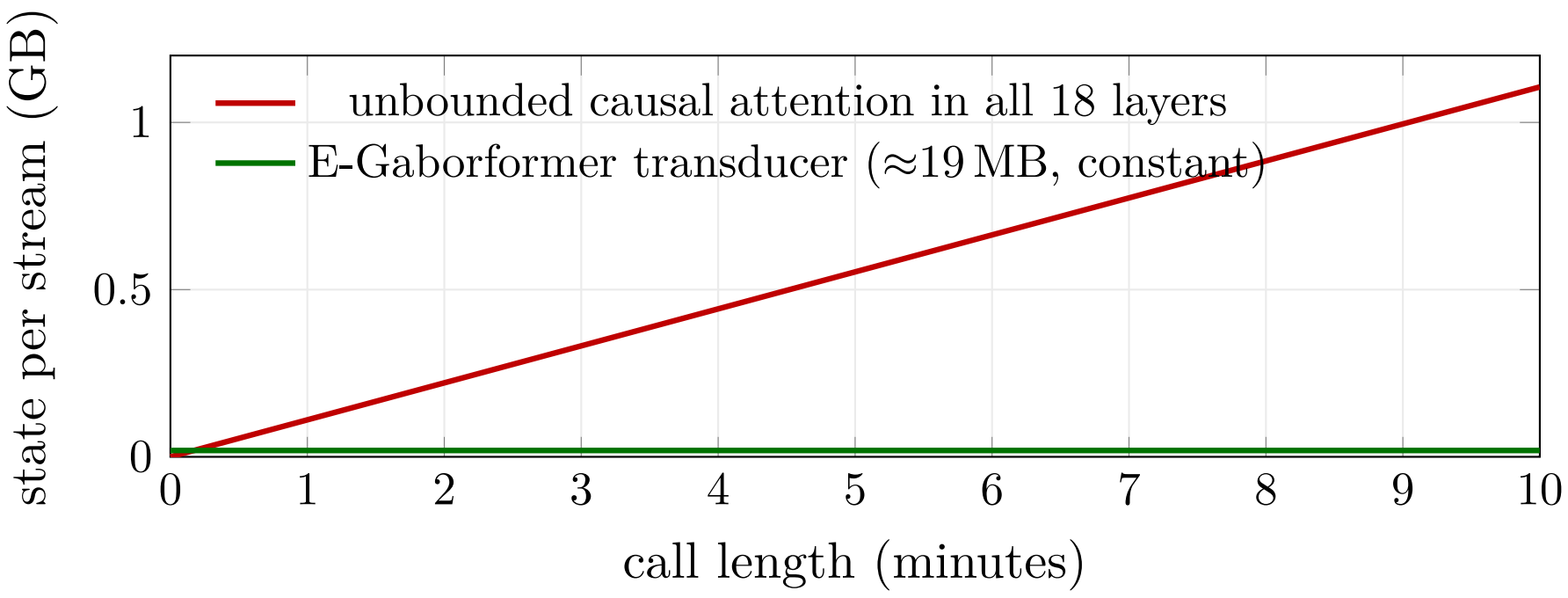
\begin{figure}[t]
\centering
\begin{tikzpicture}
\begin{axis}[width=0.72\linewidth, height=4.6cm, xmin=0, xmax=10, ymin=0, ymax=1.2,
  xlabel={call length (minutes)}, ylabel={state per stream (GB)}, grid=major, grid style={black!8},
  legend style={draw=none, fill=none, font=\small, at={(0.02,0.98)}, anchor=north west}, tick label style={font=\small}]
\addplot[red!75!black, very thick, domain=0:10] {0.1106*x};
\addlegendentry{unbounded causal attention in all 18 layers}
\addplot[green!45!black, very thick, domain=0:10] {0.019};
\addlegendentry{E-Gaborformer transducer ($\approx$19\,MB, constant)}
\end{axis}
\end{tikzpicture}
\caption{Inference state per stream at production scale ($d{=}1024$, 18 layers, fp32).}
\label{fig:mem}
\end{figure}

Because the recurrence, the windowed attention and the convolutions each look back over a fixed horizon or summarize
the past in a fixed-size state, the inference state of E-Gaborformer does not depend on how long a stream has run. Each
GDN-2 head carries $d_k d_v$ numbers, each attention layer a ring buffer of $W$ keys and values, and each convolution its
last $K{-}1$ inputs. Table~\ref{tab:state} gives the totals, and Figure~\ref{fig:mem} contrasts them with unbounded
causal attention. Unlike a pure sliding-window encoder, whose memory ends at the window, the recurrent layers also carry
context from before it.

\begin{table}[t]
\centering\small
\caption{Per-stream inference state (fp32, 80\,ms frames). Attention windows are 128 frames (10.24\,s).}
\label{tab:state}
\resizebox{\ifdim\width>\linewidth\linewidth\else\width\fi}{!}{%
\begin{tabular}{lrr}
\toprule
 & study model (12 layers, $d{=}256$) & production encoder (18 layers, $d{=}1024$) \\
\midrule
GDN-2 states & $9 \times 4$ heads: 0.6\,MB & $14 \times 16$ heads: 3.7\,MB \\
attention key/value windows & 3 layers: 0.8\,MB & 4 layers: 4.2\,MB \\
convolution caches & 1.6\,MB & 9.4\,MB \\
\midrule
total, any stream length & \textbf{$\approx$3\,MB} & \textbf{$\approx$17\,MB} (+$\approx$2\,MB prediction network) \\
unbounded causal attention after 10 min & $\approx$185\,MB & $\approx$1.1\,GB \\
\bottomrule
\end{tabular}}
\end{table}

\section{Controlled study on LibriSpeech-100}
\label{sec:study}

\subsection{Setup}
All models share one pipeline and one recipe, are trained strictly causally, and are selected and reported on the
development sets only.
\begin{itemize}
  \item \textbf{Data:} LibriSpeech train-clean-100~\citep{panayotov2015librispeech} with speed perturbation
  (0.9, 1.0, 1.1; 304\,h per epoch); dev-clean and dev-other for evaluation.
  \item \textbf{Front end:} 128 log-mel bins (25\,ms window, 10\,ms hop), SpecAugment~\citep{park2019specaugment},
  global mean--variance normalization, and a causal Conv2d stack that subsamples 8$\times$ to 80\,ms frames.
  \item \textbf{Output and loss:} a CTC output layer over 5{,}000 byte-pair-encoding (BPE) units, trained with the CTC loss
  alone (the three intermediate-CTC arms of Table~\ref{tab:explore} excepted); greedy decoding, no language model.
  \item \textbf{Baseline:} a port of ESPnet's~\citep{watanabe2018espnet} LibriSpeech-100 E-Branchformer. At its original
  front end our port has exactly ESPnet's 26{,}433{,}928 parameters; here it uses the same causal front end as
  E-Gaborformer.
  \item \textbf{Training:} Muon~\citep{jordan2024muon,liu2025muon} on hidden weight matrices and
  AdamW~\citep{loshchilov2019adamw} on all other parameters, peak learning rate $10^{-3}$ after 500 warmup updates with
  inverse square-root decay, bf16 autocast, about 2{,}330 updates per epoch on a single GPU.
  \item \textbf{E-Gaborformer:} 12 layers, $d{=}256$, global pattern GGGS repeated three times, GDN-2 with 4 heads of 64,
  attention window 128 frames, Gabor kernel size 31; 27.6M parameters against 26.8M for the E-Branchformer.
\end{itemize}
As sanity checks, every model was able to overfit a single training utterance, and we verified causality numerically:
outputs before time $t$ did not change when the input after $t$ was replaced by noise.

\subsection{E-Gaborformer versus E-Branchformer}
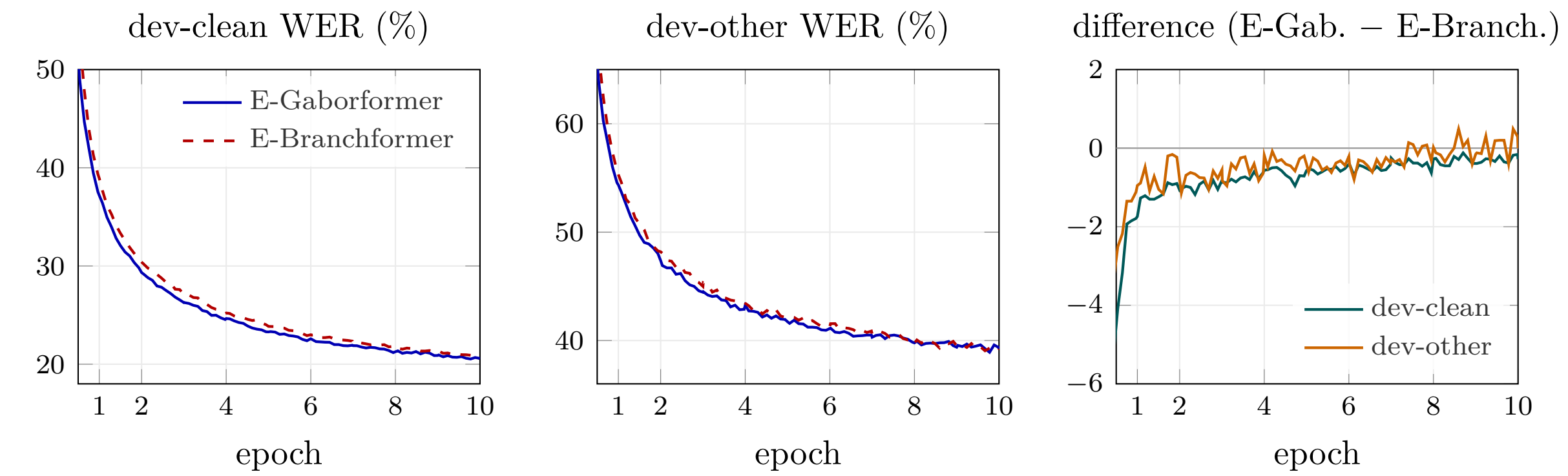
\begin{figure}[t]
\centering
\begin{tikzpicture}
\begin{groupplot}[group style={group size=3 by 1, horizontal sep=12mm},
  width=0.345\linewidth, height=4.8cm, xmin=0.5, xmax=10, xtick={1,2,4,6,8,10},
  xlabel={epoch}, grid=major, grid style={black!8}, tick label style={font=\scriptsize},
  label style={font=\small}, title style={font=\small, yshift=-1mm}, no markers]
\nextgroupplot[title={dev-clean WER (\%)}, ymin=18, ymax=50,
  legend cell align=left, legend style={draw=none, fill=white, fill opacity=0.8, font=\scriptsize, at={(0.98,0.98)}, anchor=north east}]
\addplot[blue!70!black, thick] table[x=ep, y=gc] {figs/curve10.dat}; \addlegendentry{E-Gaborformer}
\addplot[red!70!black, thick, dashed] table[x=ep, y=ec] {figs/curve10.dat}; \addlegendentry{E-Branchformer}
\nextgroupplot[title={dev-other WER (\%)}, ymin=36, ymax=65]
\addplot[blue!70!black, thick] table[x=ep, y=go] {figs/curve10.dat};
\addplot[red!70!black, thick, dashed] table[x=ep, y=eo] {figs/curve10.dat};
\nextgroupplot[title={difference (E-Gab.\ $-$ E-Branch.)}, ymin=-6, ymax=2, ytick={-6,-4,-2,0,2},
  legend cell align=left, legend style={draw=none, fill=white, fill opacity=0.8, font=\scriptsize, at={(0.98,0.03)}, anchor=south east}]
\addplot[black!40, thin, forget plot] coordinates {(0.5,0) (10,0)};
\addplot[teal!70!black, thick] table[x=ep, y=dc] {figs/curve10.dat}; \addlegendentry{dev-clean}
\addplot[orange!80!black, thick] table[x=ep, y=do] {figs/curve10.dat}; \addlegendentry{dev-other}
\end{groupplot}
\end{tikzpicture}
\caption{Development WER of causal E-Gaborformer and E-Branchformer with identical front end and recipe (mean of two
seeds, evaluated every 250 updates; greedy CTC, no language model). Right: difference between the two models; negative
values favor E-Gaborformer.}
\label{fig:curves}
\end{figure}

Figure~\ref{fig:curves} shows development-set WER over ten epochs and Table~\ref{tab:main} summarizes it. E-Gaborformer
leaves the initial blank plateau earlier and stays ahead on dev-clean for most of training: it reaches a given dev-clean
WER 5 to 18\% sooner, in updates, than the E-Branchformer. After three epochs it leads by 0.9 points on dev-clean and 0.6
on dev-other on average and is ahead in each of the three seeds (by 1.4, 0.2 and 1.2 points on dev-clean). The gap
narrows with further training. After ten epochs, both models reach 39.2\% on dev-other, and E-Gaborformer is 0.3 points
lower on dev-clean (0.35 and 0.22 in the two seeds). E-Gaborformer thus matches a causal E-Branchformer of the same size
on LibriSpeech-100 after ten epochs, with a streaming state of about 3\,MB (Section~\ref{sec:state}).

\begin{table}[t]
\centering\small
\caption{Causal models with 80\,ms frames and 128 mel bins. WER (\%) on dev-clean\,/\,dev-other, greedy CTC, no
language model. Three-epoch results are the mean of three seeds; ten-epoch results the mean of two.}
\label{tab:main}
\resizebox{\ifdim\width>\linewidth\linewidth\else\width\fi}{!}{%
\begin{tabular}{lrccc}
\toprule
 & & \multicolumn{2}{c}{3 epochs} & 10 epochs \\
\cmidrule(lr){3-4}\cmidrule(lr){5-5}
model & params & final & average of last 3 checkpoints & final \\
\midrule
E-Gaborformer (GDN-2 : attention = 3:1) & 27.6M & \textbf{\wer{26.2}{44.5}} & \textbf{\wer{25.2}{43.0}} & \textbf{\wer{20.5}{39.2}} \\
E-Branchformer & 26.8M & \wer{27.1}{45.1} & \wer{26.3}{43.7} & \wer{20.8}{39.2} \\
\bottomrule
\end{tabular}}
\end{table}

\subsection{Architecture exploration}
\label{sec:explore}
E-Gaborformer is the result of a staged architecture screen over 36 encoder configurations in about 160 training
runs, all with the recipe above and three seeds per configuration unless noted. Tables~\ref{tab:explore}
and~\ref{tab:bidir} list every configuration, with WER after one and three epochs and the number of seeds that left the
blank plateau within the first epoch. Because short runs can be dominated by how quickly each arm leaves the plateau, we
compare arms by their gap over the whole run (Table~\ref{tab:gaps}). The Gabor model overtakes unconstrained convolutions
after the first epoch and its lead grows steadily, from 0.5 points at 1.5 epochs to 1.2 at three, where it is ahead or
level in every seed (by 0.0, 1.2 and 2.5 points). E-Gaborformer is ahead of the E-Branchformer on dev-clean at every evaluation after the first, and the 3:1 hybrid
matches the 1:1 hybrid with half the attention layers, so we keep the leaner 3:1 stack. The configuration selected this
way carries over to production scale: trained as a 0.6B-parameter transducer, it has 28 to 31\% lower WER than Nemotron 3.5
ASR Streaming on English voice-agent traffic (Section~\ref{sec:prod}).

\begin{table}[t]
\centering\small
\caption{Development-clean WER difference between two arms during training (mean over seeds; negative favors the first
arm), at the same evaluation steps as Table~\ref{tab:explore}. Unconstrained convolutions lead after one epoch; the
Gabor lead then grows.}
\label{tab:gaps}
\resizebox{\linewidth}{!}{%
\begin{tabular}{lrrrrrr}
\toprule
 & \multicolumn{6}{c}{epochs (updates)} \\
arms & 1 (2{,}250) & 1.5 (3{,}500) & 2 (4{,}750) & 2.5 (5{,}750) & 3 (6{,}981) & 10 (23{,}270) \\
\midrule
Gabor vs.\ unconstrained convolutions (1:1) & $+0.4$ & $-0.5$ & $-0.7$ & $-1.0$ & $-1.2$ & --- \\
E-Gaborformer vs.\ E-Branchformer & $-1.8$ & $-1.2$ & $-1.0$ & $-1.0$ & $-0.9$ & $-0.3$ \\
3:1 vs.\ 1:1 hybrid & $-2.3$ & $-0.7$ & $-0.6$ & $-0.3$ & $0.0$ & --- \\
\bottomrule
\end{tabular}}
\end{table}

\paragraph{Round 1: bidirectional models.} We first compared the families without a causal constraint
(Table~\ref{tab:bidir}). The E-Branchformer was clearly best. A 12-layer GDN-2 encoder reached \wer{23.4}{40.8} after
three epochs against \wer{20.5}{37.1}, and every model with GDN-2 layers or Gabor convolutions had seeds that stalled on
the blank plateau; none of the three 1:1 hybrid seeds left it within an epoch. An orthonormal basis of the Gabor span
(the raw causal dictionary has a Gram condition number of about $8\times10^6$) removed most stalls in this round, but did
not improve the models that trained normally. A learning rate of $2\times10^{-3}$ left four of six architectures on the
plateau, and the Conformer did not leave it within 1{,}250 updates with AdamW alone; we therefore used Muon with a peak
learning rate of $10^{-3}$ throughout.

\paragraph{Round 2: causal models.} Under a causal constraint the ranking changed (Table~\ref{tab:explore}, top). The
Conformer left the plateau in only one of three seeds and the FastConformer only after about 2{,}000 updates. The
E-Branchformer remained the strongest non-recurrent model at \wer{28.5}{46.2}. The 1:1 hybrid with Gabor
convolutions in both the cgMLP gate and the merge reached \wer{27.4}{45.3} at three epochs, but this is the mean of the
two seeds that left the plateau; the third stalled. This is the round in which we diagnosed that problem (Section~\ref{sec:silent}).

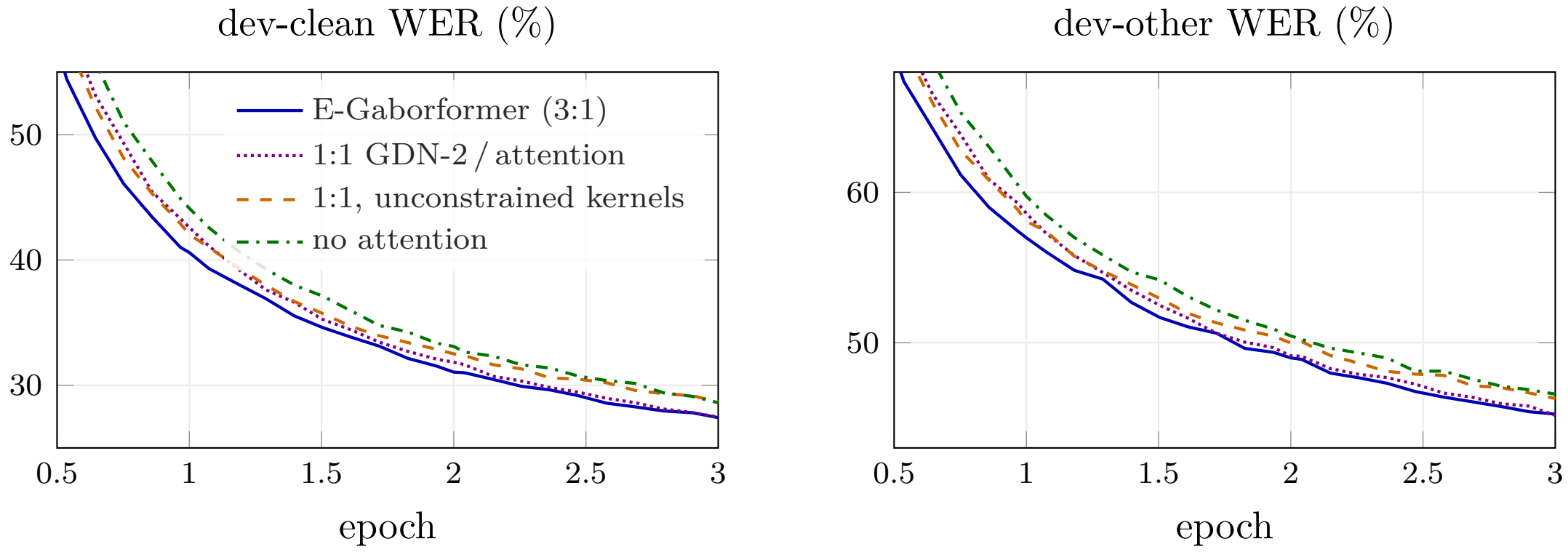
\begin{figure}[t]
\centering
\begin{tikzpicture}
\begin{groupplot}[group style={group size=2 by 1, horizontal sep=16mm},
  width=0.46\linewidth, height=5cm, xmin=0.5, xmax=3, xtick={0.5,1,1.5,2,2.5,3},
  xlabel={epoch}, grid=major, grid style={black!8}, tick label style={font=\scriptsize},
  label style={font=\small}, title style={font=\small, yshift=-1mm}, no markers]
\nextgroupplot[title={dev-clean WER (\%)}, ymin=25, ymax=55,
  legend cell align=left, legend style={draw=none, fill=white, fill opacity=0.85, font=\scriptsize, at={(0.98,0.98)}, anchor=north east}]
\addplot[blue!70!black, thick] table[x=ep, y=c] {figs/ref40.dat}; \addlegendentry{E-Gaborformer (3:1)}
\addplot[violet, thick, densely dotted] table[x=ep, y=c] {figs/one40.dat}; \addlegendentry{1:1 GDN-2\,/\,attention}
\addplot[orange!80!black, thick, dashed] table[x=ep, y=c] {figs/plain40.dat}; \addlegendentry{1:1, unconstrained kernels}
\addplot[green!45!black, thick, dashdotted] table[x=ep, y=c] {figs/noatt40.dat}; \addlegendentry{no attention}
\nextgroupplot[title={dev-other WER (\%)}, ymin=43, ymax=68]
\addplot[blue!70!black, thick] table[x=ep, y=o] {figs/ref40.dat};
\addplot[violet, thick, densely dotted] table[x=ep, y=o] {figs/one40.dat};
\addplot[orange!80!black, thick, dashed] table[x=ep, y=o] {figs/plain40.dat};
\addplot[green!45!black, thick, dashdotted] table[x=ep, y=o] {figs/noatt40.dat};
\end{groupplot}
\end{tikzpicture}
\caption{Development WER of the main round-3 configurations (mean of three seeds, 40\,ms front end).}
\label{fig:ablation}
\end{figure}

\paragraph{Round 3: sigmoid output gate.} With the sigmoid GDN-2 output gate, every seed of every configuration
trained (Table~\ref{tab:explore}, middle). Replacing both Gabor convolutions by unconstrained depthwise kernels increased
WER by 1.2 points on dev-clean and 1.1 on dev-other, and removing all attention layers increased it by 1.1 and 1.4. A 3:1
ratio of GDN-2 to attention was as accurate as 1:1 with half the attention layers. Gated attention with QK-normalization
in the attention layers~\citep{qiu2025gated} changed nothing. Intermediate CTC at layer 6 (Section~\ref{sec:prelim})
lowered dev-clean WER by 1.3 points after three epochs.
Figure~\ref{fig:ablation} shows the training curves of the main arms.

\paragraph{Round 4: front end.} To reduce encoder compute, we decreased the sequence length by adding a Conv2d stack to
the front end. This increased the frame duration from 40\,ms to 80\,ms and reduced encoder compute by nearly half, while
maintaining the accuracy of the 40\,ms model (\wer{27.4}{44.8} against \wer{27.4}{45.2}). A depthwise-separable
8$\times$ subsampler~\citep{rekesh2023fastconformer} was about one point worse. Increasing the number of mel bins from 80
to 128 lowered dev-clean WER after three epochs from 27.4 to 26.2. With 128 mel bins, adding intermediate CTC gave the
same result as the model without it (\wer{26.2}{44.5} in both cases), so the final configuration does not use it. The resulting configuration,
3:1 with Conv2d subsampling and 128 mel bins, is E-Gaborformer; the E-Branchformer with the same front end is the
baseline of Table~\ref{tab:main}.

\paragraph{Gabor convolutions across architectures.} In the causal E-Branchformer, Gabor gate convolutions raised
dev-clean WER by 2.0 points after three epochs (0.2 bidirectionally, single seed); with the SiLU output gate their effect
in the hybrids was hidden by stalled seeds; with the sigmoid gate they lowered dev-clean WER by 1.2 points in the 1:1
hybrid, the configuration from which E-Gaborformer was derived.

\begin{table}[p]
\centering\small
\caption{All causal configurations (LibriSpeech-100, greedy CTC). WER (\%) on dev-clean\,/\,dev-other after one epoch
(mean of three seeds) and three epochs (number of seeds in parentheses; in round 2 only seeds that left the plateau were
continued). ``Left plateau'': seeds below 90\% dev-clean WER within the first epoch. 1:1 and 3:1 are GDN-2\,/\,attention
ratios; ``Gabor'' and ``free'' denote Gabor-dictionary and unconstrained depthwise convolutions.}
\label{tab:explore}
\resizebox{\ifdim\width>\linewidth\linewidth\else\width\fi}{!}{%
\begin{tabular}{lccc}
\toprule
configuration & 1 epoch & left plateau & 3 epochs \\
\midrule
\multicolumn{4}{l}{\emph{Round 2: SiLU GDN-2 output gate; 40\,ms frames, 80 mel bins}} \\
Conformer & \wer{85.3}{89.4} & 1/3 & \wer{30.0}{48.0} (1) \\
FastConformer & \wer{74.5}{81.3} & 3/3 & --- \\
E-Branchformer & \wer{42.1}{58.2} & 3/3 & \wer{28.5}{46.2} (3) \\
E-Branchformer, Gabor gate conv & \wer{52.4}{66.1} & 3/3 & \wer{30.6}{48.5} (3) \\
GDN-2 only, Gabor gate conv & \wer{66.4}{76.4} & 2/3 & \wer{29.8}{47.4} (2) \\
GDN-2 only, free convs & \wer{51.7}{65.7} & 3/3 & --- \\
1:1, Gabor gate conv & \wer{67.7}{77.0} & 2/3 & \wer{30.4}{48.1} (2) \\
1:1, gate conv half Gabor, half free & \wer{47.9}{62.7} & 3/3 & \wer{29.8}{47.3} (3) \\
1:1, free gate conv & \wer{64.9}{75.1} & 2/3 & \wer{29.2}{47.1} (2) \\
1:1, Gabor gate and merge convs & \wer{61.1}{71.9} & 2/3 & \wer{27.4}{45.3} (2) \\
\midrule
\multicolumn{4}{l}{\emph{Round 3: sigmoid output gate, RMSNorm $\epsilon = 10^{-6}$; 40\,ms frames, 80 mel bins}} \\
GDN-2 only, Gabor gate conv & \wer{44.9}{60.5} & 3/3 & \wer{28.6}{46.6} (3) \\
1:1, Gabor gate and merge convs & \wer{43.3}{59.3} & 3/3 & \wer{27.5}{45.2} (3) \\
1:1, free gate and merge convs & \wer{42.9}{58.9} & 3/3 & \wer{28.7}{46.3} (3) \\
3:1, Gabor gate and merge convs & \wer{41.0}{57.4} & 3/3 & \wer{27.4}{45.2} (3) \\
1:1, gated attention with QK-norm & \wer{41.5}{57.9} & 3/3 & \wer{27.4}{45.2} (3) \\
1:1, intermediate CTC at layer 6 & \wer{41.5}{57.9} & 3/3 & \wer{26.2}{43.9} (3) \\
1:1, self-conditioned intermediate CTC & \wer{42.6}{58.6} & 3/3 & \wer{26.4}{44.4} (3) \\
\midrule
\multicolumn{4}{l}{\emph{Round 4: 3:1, attention window 128, 80\,ms frames}} \\
Conv2d 8$\times$, 80 mel & \wer{41.6}{57.7} & 3/3 & \wer{27.4}{44.8} (3) \\
depthwise-separable 8$\times$, 80 mel & \wer{46.6}{61.6} & 3/3 & \wer{28.7}{46.6} (3) \\
\textbf{Conv2d 8$\times$, 128 mel (E-Gaborformer)} & \textbf{\wer{37.4}{54.6}} & 3/3 & \textbf{\wer{26.2}{44.5}} (3) \\
depthwise-separable 8$\times$, 128 mel & \wer{39.9}{56.9} & 3/3 & \wer{27.1}{45.6} (3) \\
Conv2d 8$\times$, 128 mel, intermediate CTC & \wer{37.6}{54.7} & 3/3 & \wer{26.2}{44.5} (3) \\
E-Branchformer, Conv2d 8$\times$, 128 mel & \wer{39.2}{55.5} & 3/3 & \wer{27.1}{45.1} (3) \\
\bottomrule
\end{tabular}}
\end{table}

\begin{table}[t]
\centering\small
\caption{Bidirectional configurations (round 1; SiLU GDN-2 output gate, 40\,ms frames, 80 mel bins). WER (\%) on
dev-clean\,/\,dev-other after one epoch (three seeds; seeds that left the plateau in parentheses) with the raw and the
orthonormalized Gabor dictionary, and after three epochs (number of seeds).}
\label{tab:bidir}
\resizebox{\ifdim\width>\linewidth\linewidth\else\width\fi}{!}{%
\begin{tabular}{lccc}
\toprule
configuration & 1 epoch, raw dictionary & 1 epoch, orthonormal basis & 3 epochs \\
\midrule
Conformer & \wer{46.4}{60.6} (3) & --- & --- \\
E-Branchformer & \textbf{\wer{32.6}{48.7}} (3) & --- & \textbf{\wer{20.5}{37.1}} (3) \\
E-Branchformer, Gabor gate conv & \wer{77.7}{83.1} (1) & \wer{33.9}{50.0} (3) & \wer{20.8}{37.6} (1) \\
GDN-2 only, Gabor gate conv & \wer{79.5}{85.0} (1) & \wer{65.9}{75.6} (2) & \wer{23.4}{40.8} (1) \\
1:1, Gabor gate conv & \wer{99.6}{99.8} (0) & \wer{59.1}{69.6} (2) & \wer{24.4}{41.0} (1, orthonormal) \\
1:1, gate conv half Gabor, half free & \wer{52.4}{64.1} (2) & \wer{54.6}{66.3} (3) & \wer{20.8}{37.6} (2) \\
1:1, free gate conv & \wer{58.9}{69.8} (2) & --- & --- \\
1:1, Gabor gate and merge convs & \wer{56.8}{67.8} (2) & \wer{43.6}{58.8} (3) & \wer{21.7}{38.5} (2) \\
\bottomrule
\end{tabular}}
\end{table}

\paragraph{Causal and bidirectional models.} The two settings rank the architectures differently. Bidirectionally, the
GDN-2 encoder trailed the E-Branchformer by 2.9 points on dev-clean after three epochs; causally, the hybrids matched or
led it, which suggests that recurrence is especially well matched to the causal setting that streaming requires.

\section{A streaming transducer at scale}
\label{sec:prod}

Once the architecture was settled on LibriSpeech, we trained it at the scale of the recognizer behind our real-time voice
agents: Hebrew and English, telephone-quality and spontaneous speech, strictly causal or low-lookahead decoding.

\subsection{Model}
\begin{itemize}
  \item \textbf{Encoder:} 18 E-Gaborformer layers, $d{=}1024$, feed-forward and cgMLP width 4096 (Gabor convolutions over
  2048 channels, kernel 31, in both the gate and the merge), 128 mel bins, Conv2d 8$\times$ subsampling to 80\,ms frames.
  The global stack follows the 3:1 pattern: 14 GDN-2 layers (16 heads of 64) and 4 sliding-window attention layers
  (16 heads, window 128 frames). The encoder has 598M parameters.
  \item \textbf{Transducer:} an RNN-T~\citep{graves2012rnnt} with a 640-dimensional joint network over a 13{,}100-piece
  SentencePiece vocabulary plus blank. The vocabulary extends the 13{,}087-piece vocabulary of Nemotron 3.5 ASR
  Streaming~\citep{nemotron2026asr} with symbols needed for written-form output (for example \texttt{\%}, \texttt{:} and
  currency signs).
  \item \textbf{Hybrid prediction network} (Section~\ref{sec:pred}): four layers alternating GDN-2 and causal token
  attention, 17.9M parameters including the embedding.
\end{itemize}
The full model has 626M parameters, about the size of Nemotron 3.5 ASR Streaming (0.6B).

\subsection{Training}
Training used about 12k hours of Hebrew and 96k hours of English, with Hebrew upweighted in sampling, speed perturbation
and SpecAugment, on three or four H200 GPUs with Muon and AdamW in bf16, with roughly 8{,}500\,s of audio per update. The model with a
two-layer GDN-2 prediction network was trained for 115k updates, and the hybrid prediction network of
Section~\ref{sec:pred} for a further 44k. To shorten training, we used Nemotron 3.5 ASR Streaming as a teacher during
the first 10{,}750 updates: its token embedding initialized ours, and two distillation terms were added to the RNN-T loss,
one matching our encoder output to Nemotron's (through a learned projection) and one matching the transducer's
blank\,/\,next-token\,/\,other probabilities on the reference alignment lattice~\citep{panchapagesan2021efficient}. The architecture itself was fixed
beforehand by the teacher-free study of Section~\ref{sec:study}, and all later training used the RNN-T loss alone.

\subsection{A hybrid prediction network}
\label{sec:pred}
We expected a purely recurrent prediction network to have difficulty copying recent tokens such as digits, names and
parts of an address, and therefore apply the encoder's design to the prediction network: it alternates GDN-2 layers (10 heads of 64) with
causal \emph{token attention} over the last 128 tokens (rotary positions, RMS-normalized queries and keys), each in a
pre-norm residual block. For streaming, each attention layer keeps a ring buffer of 128 keys and values and a position
counter per stream. The state is therefore of fixed size and can be selected per row in batched frame-synchronous
greedy decoding, exactly like the recurrent state. Step-by-step decoding matches the full-sequence computation to
numerical precision, including after the ring buffer wraps.

The hybrid network replaced a two-layer GDN-2 prediction network in an already trained model. Its first and third
layers were copied from the two trained GDN-2 layers and its attention layers initialized randomly. With the encoder and
the token embedding frozen, the new network was first distilled~\citep{hinton2015distilling} from the previous model by
minimizing the full-vocabulary KL divergence from the previous model's output distribution to the new one's at every
lattice node. Training then continued with the RNN-T loss, the
encoder unfrozen at a reduced learning rate, and optimizer states carried over by parameter name.

\subsection{Selectable lookahead}
\label{sec:la}
Following cache-aware streaming~\citep{noroozi2024stateful}, the sliding-window attention layers use a chunked mask:
query $i$ attends to keys $j$ with $i - W < j \le e(i)$, where $e(i)$ is the last frame of the chunk of size $C$ that
contains $i$. This gives at most $C{-}1$ frames of lookahead. Because the recurrent layers and the convolutions remain
strictly causal, the bound is the same in every layer and does not accumulate with depth. During training, $C$ is drawn
per micro-batch from $\{1, 2, 7, 14\}$ (0, 80, 480 and 1040\,ms of lookahead). At inference, the same weights run
strictly causally ($C{=}1$) or with up to 480\,ms of lookahead ($C{=}7$).

\subsection{Results}
\label{sec:results}
Table~\ref{tab:prod} compares the current transducer with our previous production E-Gaborformer transducer. The test
sets are held-out samples of the speech our
voice agents receive in production: general and hard (noisy, spontaneous) Hebrew and English, Hebrew with many personal
names, spoken addresses, URLs and email addresses, and real WhatsApp voice messages, all from our internal evaluation suite. All systems are decoded greedily without an
external language model, and all hypotheses and references are normalized identically.

\begin{table}[t]
\centering\small
\caption{Production test sets, WER (\%), greedy decoding, no external language model. Lookahead 0\,ms unless stated.
WhatsApp: 54 Hebrew voice messages (1.18\,h, 8{,}941 words) decoded in 20\,s segments.}
\label{tab:prod}
\resizebox{\ifdim\width>\linewidth\linewidth\else\width\fi}{!}{%
\begin{tabular}{lccr}
\toprule
test set & previous production model & current model & relative \\
\midrule
Hebrew, general & 12.3 & \textbf{10.9} & $-11\%$ \\
Hebrew, hard & 14.2 & \textbf{11.5} & $-19\%$ \\
Hebrew, names & 10.2 & \textbf{8.2} & $-20\%$ \\
Hebrew, addresses, URLs and emails & \textbf{14.8} & 15.8 & $+7\%$ \\
English, general & \textbf{9.5} & 9.6 & $+1\%$ \\
English, hard & 11.0 & 11.0 & $0\%$ \\
\midrule
WhatsApp voice messages & 16.4 & 16.0 & $-2\%$ \\
WhatsApp voice messages, 480\,ms lookahead & 16.3 & 16.0 & $-2\%$ \\
\bottomrule
\end{tabular}}
\end{table}

Compared with our previous production transducer, the current model lowers Hebrew WER by 11 to 20\% relative on three of
the four Hebrew sets, with the largest gains on the hardest and most name-heavy sets, while holding English at the same level. On real voice messages,
WER is identical with and without 480\,ms of lookahead (16.0\%): the model can be served strictly causally at no cost in
accuracy, because only 4 of its 18 layers can see future frames at all.

\paragraph{Comparison with Nemotron 3.5 ASR Streaming.} On the general English set (real production speech, 24
recordings, 1{,}214 reference words after normalization) we also decoded NVIDIA's Nemotron 3.5 ASR
Streaming~\citep{nemotron2026asr}, a public cache-aware transducer of the same size, with the English prompt, identical
audio handling and identical text normalization (Table~\ref{tab:nemotron}). Its attention context was set to the
streaming configurations [56,\,0] and [56,\,6] (left and right context in frames), matching our causal mode and our
480\,ms lookahead. E-Gaborformer has 31\% lower WER than Nemotron when both run causally and 28\% lower with 480\,ms of
lookahead. Lookahead helps Nemotron by 0.7 points, while E-Gaborformer is equally accurate without it. The comparison reflects the accuracy a voice agent obtains from each model on this traffic, for
which our model was trained (Nemotron also served as an early teacher, Section~\ref{sec:prod}).

\begin{table}[t]
\centering\small
\caption{General English production set (24 recordings, 1{,}214 words), WER (\%), greedy decoding, identical
normalization.}
\label{tab:nemotron}
\begin{tabular}{lrcc}
\toprule
model & params & causal & 480\,ms lookahead \\
\midrule
Nemotron 3.5 ASR Streaming & 0.6B & 14.0 & 13.3 \\
E-Gaborformer transducer & 0.6B & \textbf{9.6} & \textbf{9.6} \\
\bottomrule
\end{tabular}
\end{table}

\subsection{Training efficiency}
\label{sec:speed}
Profiling a training update of the transducer showed that most of the time was spent on the full-vocabulary
log-softmax, although the RNN-T loss uses only two entries per lattice node. The changes in Table~\ref{tab:speed}
reduce the time per update by 2.6$\times$ relative to torchaudio's RNN-T loss with an fp32 joint network, without
changing the batch; all keep the loss and gradients exact except the bf16 joint matrix multiplications.

\paragraph{Two-class RNN-T lattice.} The RNN-T recursion needs only two log-probabilities per lattice node $(t,u)$: emit
the next token $y_{u+1}$, or emit blank. Standard implementations nevertheless run their dynamic program over the full
$(T, U{+}1, V{+}1)$ log-softmax. We compute the joint logits $z_{t,u}$ in chunks, take
$\ell^{y}_{t,u} = z_{t,u}[y_{u+1}] - \mathrm{lse}(z_{t,u})$ and $\ell^{b}_{t,u} = z_{t,u}[\varnothing] - \mathrm{lse}(z_{t,u})$,
where $\mathrm{lse}$ is the log-sum-exp over the vocabulary and $\varnothing$ the blank, and run the forward-backward
recursion on the two-class lattice. Its gradients $\gamma^{y} = \partial\mathcal{L}/\partial\ell^{y}$ and
$\gamma^{b} = \partial\mathcal{L}/\partial\ell^{b}$ give the logit gradient in closed form, with $\mathbf{e}$ a one-hot
vector,
\begin{equation}
  \frac{\partial \mathcal{L}}{\partial z_{t,u}} = \gamma^{y}_{t,u}\, \mathbf{e}_{y_{u+1}} + \gamma^{b}_{t,u}\, \mathbf{e}_{\varnothing}
  - \bigl(\gamma^{y}_{t,u} + \gamma^{b}_{t,u}\bigr)\, \mathrm{softmax}(z_{t,u}),
\end{equation}
which is propagated through the joint network by matrix multiplication. The loss and gradients are identical to those
of the full lattice, and the vocabulary-wide work reduces to one logsumexp pass and one gradient pass. The approach is
related to function merging~\citep{li2019rnnt}; unlike pruned RNN-T~\citep{kuang2022pruned}, it keeps every path.

\paragraph{Aligned output layer.} With 13{,}101 outputs, the joint network's matrix multiplications fell back to slower
GPU kernels. Padding the output layer to a multiple of 128 rows, with zero weights and a bias of $-10^4$ so that the
padding contributes nothing to the softmax, restores the tensor-core kernels.

\paragraph{Toeplitz-form depthwise convolutions.} A depthwise convolution with kernel 31 performs little arithmetic, but
generic implementations issue one unaligned load per tap and output. We write each channel's convolution over a block
of the batch as a product with a Toeplitz matrix built from its kernel and evaluate it on tensor cores. The input
gradient is the same operation with the kernel reversed, and the weight gradient is a small matrix product followed by
diagonal sums. For a $(30, 2048, 300)$ bf16 input, the forward and backward passes drop from 2.1\,ms to 0.54\,ms.

\paragraph{Length regrouping with preserved weights.} Micro-batches are formed by a token budget and were on average
only 61\% real frames. We pool the utterances of all micro-batches of one update, sort them by length and refill the
micro-batches, which raises the fill to 80--83\%. Each utterance keeps the loss weight it had in its original
micro-batch, so the update is unchanged; a unit test verifies equality to $10^{-4}$.

With live data on three GPUs, new tensor shapes also triggered per-shape setup work and host--device synchronizations
stalled the GPUs. Padding lengths to multiples of 1.28\,s, running the subsampling convolutions in fixed batch chunks,
computing sequence lengths on the host and preparing the next update in a background thread removed these stalls.

\begin{table}[t]
\centering\small
\caption{Training time per update (same batch and loss). Single-GPU rows replay the same eight real micro-batches
(about 2{,}765\,s of audio per update); the last rows are live training on three GPUs.}
\label{tab:speed}
\resizebox{\ifdim\width>\linewidth\linewidth\else\width\fi}{!}{%
\begin{tabular}{lr}
\toprule
configuration & seconds / update \\
\midrule
baseline (full-vocabulary RNN-T lattice, fp32 joint) & 7.55 \\
+ two-class lattice with closed-form gradient & 6.02 \\
+ compiled lattice passes & 5.27 \\
+ output layer aligned to 128 rows, bf16 joint matrix multiplications & 3.75 \\
+ Toeplitz-form depthwise convolutions & 3.33 \\
+ length regrouping with preserved weights & 2.92 \\
+ channels-last subsampling, batched Muon & \textbf{2.90} (2.6$\times$) \\
\midrule
three GPUs, live data: baseline & 9.0 \\
three GPUs, live data: all changes, stable shapes, prefetching & \textbf{3.58} (2.5$\times$) \\
\bottomrule
\end{tabular}}
\end{table}

\section{Conclusion}
E-Gaborformer shows that a mostly recurrent, strictly causal encoder is a strong foundation for real-time speech
recognition. On LibriSpeech-100 it matched or beat a causal E-Branchformer of the same size
(\wer{20.5}{39.2} against \wer{20.8}{39.2} after ten epochs) and reached a given WER up to 18\% sooner, with a constant
streaming state of about 3\,MB. Three ingredients made this work: a few attention layers for exact recall, a GDN-2
output gate that keeps the recurrent branch active from the first update (0 of 36 runs stalled, against 4 of 18), and
Gabor-dictionary convolutions, which led unconstrained ones by 1.2 points at three epochs in the hybrid encoder. The design extends naturally
to the transducer's prediction network and scales to a 0.6B-parameter Hebrew and English recognizer that serves several
lookaheads from one set of weights, keeps a constant 19\,MB state per call, trains 2.6$\times$ faster with an exact
objective, and has 28 to 31\% lower WER than Nemotron 3.5 ASR Streaming on English voice-agent traffic.

Natural next steps are longer controlled training runs, emission-latency measurements, and evaluation on multi-minute
streams, where a constant-size state is most valuable.

\bibliographystyle{plainnat}
\bibliography{refs}

\begin{thebibliography}{57}
\providecommand{\natexlab}[1]{#1}
\providecommand{\url}[1]{\texttt{#1}}
\expandafter\ifx\csname urlstyle\endcsname\relax
  \providecommand{\doi}[1]{doi: #1}\else
  \providecommand{\doi}{doi: \begingroup \urlstyle{rm}\Url}\fi

\bibitem[Alekseev and Bobe(2019)]{alekseev2019gabornet}
Andrey Alekseev and Anatoly Bobe.
\newblock {GaborNet}: Gabor filters with learnable parameters in deep
  convolutional neural network.
\newblock In \emph{Proc. International Conference on Engineering and
  Telecommunication (EnT)}, 2019.

\bibitem[Chang and Morgan(2014)]{chang2014robust}
Shuo-Yiin Chang and Nelson Morgan.
\newblock Robust {CNN}-based speech recognition with {Gabor} filter kernels.
\newblock In \emph{Proc. Interspeech}, 2014.

\bibitem[Chen et~al.(2021)Chen, Wu, Wang, Liu, and Li]{chen2021streaming}
Xie Chen, Yu~Wu, Zhenghao Wang, Shujie Liu, and Jinyu Li.
\newblock Developing real-time streaming transformer transducer for speech
  recognition on large-scale dataset.
\newblock In \emph{Proc. ICASSP}, 2021.

\bibitem[De et~al.(2024)De, Smith, Fernando, Botev, et~al.]{de2024griffin}
Soham De, Samuel~L. Smith, Anushan Fernando, Aleksandar Botev, et~al.
\newblock Griffin: Mixing gated linear recurrences with local attention for
  efficient language models.
\newblock \emph{arXiv preprint arXiv:2402.19427}, 2024.

\bibitem[Ding et~al.(2017)Ding, Patel, Chen, Butler, Luo, and
  Poeppel]{ding2017temporal}
Nai Ding, Aniruddh~D. Patel, Lin Chen, Henry Butler, Cheng Luo, and David
  Poeppel.
\newblock Temporal modulations in speech and music.
\newblock \emph{Neuroscience \& Biobehavioral Reviews}, 81:\penalty0 181--187,
  2017.

\bibitem[Gabor(1946)]{gabor1946theory}
Dennis Gabor.
\newblock Theory of communication.
\newblock \emph{Journal of the Institution of Electrical Engineers},
  93\penalty0 (26):\penalty0 429--457, 1946.

\bibitem[Graves(2012)]{graves2012rnnt}
Alex Graves.
\newblock Sequence transduction with recurrent neural networks.
\newblock \emph{arXiv preprint arXiv:1211.3711}, 2012.

\bibitem[Graves et~al.(2006)Graves, Fern{\'a}ndez, Gomez, and
  Schmidhuber]{graves2006ctc}
Alex Graves, Santiago Fern{\'a}ndez, Faustino Gomez, and J{\"u}rgen
  Schmidhuber.
\newblock Connectionist temporal classification: labelling unsegmented sequence
  data with recurrent neural networks.
\newblock In \emph{Proc. ICML}, 2006.

\bibitem[Gu and Dao(2024)]{gu2024mamba}
Albert Gu and Tri Dao.
\newblock Mamba: Linear-time sequence modeling with selective state spaces.
\newblock In \emph{Proc. COLM}, 2024.

\bibitem[Gu et~al.(2022)Gu, Goel, and R{\'e}]{gu2022s4}
Albert Gu, Karan Goel, and Christopher R{\'e}.
\newblock Efficiently modeling long sequences with structured state spaces.
\newblock In \emph{Proc. ICLR}, 2022.

\bibitem[Gulati et~al.(2020)Gulati, Qin, Chiu, Parmar, Zhang, Yu, Han, Wang,
  Zhang, Wu, and Pang]{gulati2020conformer}
Anmol Gulati, James Qin, Chung-Cheng Chiu, Niki Parmar, Yu~Zhang, Jiahui Yu,
  Wei Han, Shibo Wang, Zhengdong Zhang, Yonghui Wu, and Ruoming Pang.
\newblock Conformer: Convolution-augmented transformer for speech recognition.
\newblock In \emph{Proc. Interspeech}, 2020.

\bibitem[Hatamizadeh et~al.(2026)Hatamizadeh, Choi, and
  Kautz]{hatamizadeh2026gdn2}
Ali Hatamizadeh, Yejin Choi, and Jan Kautz.
\newblock Gated {DeltaNet-2}: Decoupling erase and write in linear attention.
\newblock \emph{arXiv preprint arXiv:2605.22791}, 2026.

\bibitem[He et~al.(2019)He, Sainath, Prabhavalkar, et~al.]{he2019streaming}
Yanzhang He, Tara~N. Sainath, Rohit Prabhavalkar, et~al.
\newblock Streaming end-to-end speech recognition for mobile devices.
\newblock In \emph{Proc. ICASSP}, 2019.

\bibitem[Hinton et~al.(2015)Hinton, Vinyals, and Dean]{hinton2015distilling}
Geoffrey Hinton, Oriol Vinyals, and Jeff Dean.
\newblock Distilling the knowledge in a neural network.
\newblock \emph{arXiv preprint arXiv:1503.02531}, 2015.

\bibitem[Jacobsen et~al.(2016)Jacobsen, van Gemert, Lou, and
  Smeulders]{jacobsen2016structured}
J{\"o}rn-Henrik Jacobsen, Jan van Gemert, Zhongyu Lou, and Arnold W.~M.
  Smeulders.
\newblock Structured receptive fields in {CNNs}.
\newblock In \emph{Proc. CVPR}, 2016.

\bibitem[Jiang et~al.(2024)Jiang, Li, Florea, Han, and
  Mesgarani]{jiang2024speechslytherin}
Xilin Jiang, Yinghao~Aaron Li, Adrian~Nicolas Florea, Cong Han, and Nima
  Mesgarani.
\newblock Speech {S}lytherin: Examining the performance and efficiency of
  {M}amba for speech separation, recognition, and synthesis.
\newblock \emph{arXiv preprint arXiv:2407.09732}, 2024.

\bibitem[Jordan et~al.(2024)Jordan, Jin, Boza, You, Cesista, Newhouse, and
  Bernstein]{jordan2024muon}
Keller Jordan, Yuchen Jin, Vlado Boza, Jiacheng You, Franz Cesista, Laker
  Newhouse, and Jeremy Bernstein.
\newblock Muon: An optimizer for hidden layers in neural networks.
\newblock \url{https://kellerjordan.github.io/posts/muon/}, 2024.

\bibitem[Katharopoulos et~al.(2020)Katharopoulos, Vyas, Pappas, and
  Fleuret]{katharopoulos2020transformers}
Angelos Katharopoulos, Apoorv Vyas, Nikolaos Pappas, and Fran{\c{c}}ois
  Fleuret.
\newblock Transformers are {RNNs}: Fast autoregressive transformers with linear
  attention.
\newblock In \emph{Proc. ICML}, 2020.

\bibitem[Kim et~al.(2023)Kim, Wu, Peng, Pan, Sridhar, Han, and
  Watanabe]{kim2023ebranchformer}
Kwangyoun Kim, Felix Wu, Yifan Peng, Jing Pan, Prashant Sridhar, Kyu~J. Han,
  and Shinji Watanabe.
\newblock {E-Branchformer}: Branchformer with enhanced merging for speech
  recognition.
\newblock In \emph{Proc. IEEE SLT}, 2023.

\bibitem[Kuang et~al.(2022)Kuang, Guo, Kang, Lin, Luo, Yao, and
  Povey]{kuang2022pruned}
Fangjun Kuang, Liyong Guo, Wei Kang, Long Lin, Mingshuang Luo, Zengwei Yao, and
  Daniel Povey.
\newblock Pruned {RNN-T} for fast, memory-efficient {ASR} training.
\newblock In \emph{Proc. Interspeech}, 2022.

\bibitem[Lee and Watanabe(2021)]{lee2021interctc}
Jaesong Lee and Shinji Watanabe.
\newblock Intermediate loss regularization for {CTC}-based speech recognition.
\newblock In \emph{Proc. ICASSP}, 2021.

\bibitem[Li et~al.(2019)Li, Zhao, Hu, and Gong]{li2019rnnt}
Jinyu Li, Rui Zhao, Hu~Hu, and Yifan Gong.
\newblock Improving {RNN} transducer modeling for end-to-end speech
  recognition.
\newblock In \emph{Proc. IEEE ASRU}, 2019.

\bibitem[Lieber et~al.(2024)]{lieber2024jamba}
Opher Lieber et~al.
\newblock Jamba: A hybrid {Transformer-Mamba} language model.
\newblock \emph{arXiv preprint arXiv:2403.19887}, 2024.

\bibitem[Liu et~al.(2025)]{liu2025muon}
Jingyuan Liu et~al.
\newblock Muon is scalable for {LLM} training.
\newblock \emph{arXiv preprint arXiv:2502.16982}, 2025.

\bibitem[Loshchilov and Hutter(2019)]{loshchilov2019adamw}
Ilya Loshchilov and Frank Hutter.
\newblock Decoupled weight decay regularization.
\newblock In \emph{Proc. ICLR}, 2019.

\bibitem[Luan et~al.(2018)Luan, Chen, Zhang, Han, and Liu]{luan2018gabor}
Shangzhen Luan, Chen Chen, Baochang Zhang, Jungong Han, and Jianzhuang Liu.
\newblock Gabor convolutional networks.
\newblock \emph{IEEE Transactions on Image Processing}, 27\penalty0
  (9):\penalty0 4357--4366, 2018.

\bibitem[Noroozi et~al.(2024)Noroozi, Majumdar, Kumar, Balam, and
  Ginsburg]{noroozi2024stateful}
Vahid Noroozi, Somshubra Majumdar, Ankur Kumar, Jagadeesh Balam, and Boris
  Ginsburg.
\newblock Stateful {Conformer} with cache-based inference for streaming
  automatic speech recognition.
\newblock In \emph{Proc. ICASSP}, 2024.

\bibitem[Nozaki and Komatsu(2021)]{nozaki2021selfcond}
Jumon Nozaki and Tatsuya Komatsu.
\newblock Relaxing the conditional independence assumption of {CTC}-based {ASR}
  by conditioning on intermediate predictions.
\newblock In \emph{Proc. Interspeech}, 2021.

\bibitem[{NVIDIA}(2026)]{nemotron2026asr}
{NVIDIA}.
\newblock Nemotron 3.5 {ASR} streaming 0.6{B}.
\newblock \url{https://huggingface.co/nvidia/nemotron-3.5-asr-streaming-0.6b},
  2026.

\bibitem[Panayotov et~al.(2015)Panayotov, Chen, Povey, and
  Khudanpur]{panayotov2015librispeech}
Vassil Panayotov, Guoguo Chen, Daniel Povey, and Sanjeev Khudanpur.
\newblock {LibriSpeech}: an {ASR} corpus based on public domain audio books.
\newblock In \emph{Proc. ICASSP}, 2015.

\bibitem[Panchapagesan et~al.(2021)Panchapagesan, Park, Chiu, Shangguan, Liang,
  and Gruenstein]{panchapagesan2021efficient}
Sankaran Panchapagesan, Daniel~S. Park, Chung-Cheng Chiu, Yuan Shangguan, Qiao
  Liang, and Alexander Gruenstein.
\newblock Efficient knowledge distillation for {RNN}-transducer models.
\newblock In \emph{Proc. ICASSP}, 2021.

\bibitem[Park et~al.(2019)Park, Chan, Zhang, Chiu, Zoph, Cubuk, and
  Le]{park2019specaugment}
Daniel~S. Park, William Chan, Yu~Zhang, Chung-Cheng Chiu, Barret Zoph, Ekin~D.
  Cubuk, and Quoc~V. Le.
\newblock {SpecAugment}: A simple data augmentation method for automatic speech
  recognition.
\newblock In \emph{Proc. Interspeech}, 2019.

\bibitem[Peng et~al.(2022)Peng, Dalmia, Lane, and
  Watanabe]{peng2022branchformer}
Yifan Peng, Siddharth Dalmia, Ian Lane, and Shinji Watanabe.
\newblock Branchformer: Parallel {MLP}-attention architectures to capture local
  and global context for speech recognition and understanding.
\newblock In \emph{Proc. ICML}, 2022.

\bibitem[Peng et~al.(2023)Peng, Kim, Wu, Yan, Arora, Chen, Tang, Shon, Sridhar,
  and Watanabe]{peng2023comparative}
Yifan Peng, Kwangyoun Kim, Felix Wu, Brian Yan, Siddhant Arora, William Chen,
  Jiyang Tang, Suwon Shon, Prashant Sridhar, and Shinji Watanabe.
\newblock A comparative study on {E-Branchformer} vs {Conformer} in speech
  recognition, translation, and understanding tasks.
\newblock In \emph{Proc. Interspeech}, 2023.

\bibitem[Poli et~al.(2023)Poli, Massaroli, Nguyen, et~al.]{poli2023hyena}
Michael Poli, Stefano Massaroli, Eric Nguyen, et~al.
\newblock Hyena hierarchy: Towards larger convolutional language models.
\newblock In \emph{Proc. ICML}, 2023.

\bibitem[Qiu et~al.(2018)Qiu, Cheng, Calderbank, and Sapiro]{qiu2018dcfnet}
Qiang Qiu, Xiuyuan Cheng, Robert Calderbank, and Guillermo Sapiro.
\newblock {DCFNet}: Deep neural network with decomposed convolutional filters.
\newblock In \emph{Proc. ICML}, 2018.

\bibitem[Qiu et~al.(2025)]{qiu2025gated}
Zihan Qiu et~al.
\newblock Gated attention for large language models: Non-linearity, sparsity,
  and attention-sink-free.
\newblock \emph{arXiv preprint arXiv:2505.06708}, 2025.

\bibitem[Ravanelli and Bengio(2018)]{ravanelli2018sincnet}
Mirco Ravanelli and Yoshua Bengio.
\newblock Speaker recognition from raw waveform with {SincNet}.
\newblock In \emph{Proc. IEEE SLT}, 2018.

\bibitem[Rekesh et~al.(2023)Rekesh, Koluguri, Kriman, Majumdar, Noroozi, Huang,
  Hrinchuk, Puvvada, Kumar, Balam, and Ginsburg]{rekesh2023fastconformer}
Dima Rekesh, Nithin~Rao Koluguri, Samuel Kriman, Somshubra Majumdar, Vahid
  Noroozi, He~Huang, Oleksii Hrinchuk, Krishna Puvvada, Ankur Kumar, Jagadeesh
  Balam, and Boris Ginsburg.
\newblock Fast {Conformer} with linearly scalable attention for efficient
  speech recognition.
\newblock In \emph{Proc. IEEE ASRU}, 2023.

\bibitem[Ren et~al.(2024)Ren, Liu, Lu, Shen, Liang, and Chen]{ren2024samba}
Liliang Ren, Yang Liu, Yadong Lu, Yelong Shen, Chen Liang, and Weizhu Chen.
\newblock Samba: Simple hybrid state space models for efficient unlimited
  context language modeling.
\newblock \emph{arXiv preprint arXiv:2406.07522}, 2024.

\bibitem[Roll et~al.(2026)Roll, Yi, Mar{\c{s}}an, Grenez, Stein, Mrkaic,
  Padjin, Zeljkovic, and Graham]{orukeet2026}
Nathan Roll, Irene Yi, B{\"u}{\c{s}}ra Mar{\c{s}}an, Vianney Grenez, Gabriel
  Stein, Momcilo Mrkaic, Pavle Padjin, Vladimir Zeljkovic, and Calbert Graham.
\newblock Orukeet: Multilingual {ASR} with frozen {Gabor} kernels.
\newblock \emph{arXiv preprint arXiv:2609.10054}, 2026.

\bibitem[Sch{\"a}dler et~al.(2012)Sch{\"a}dler, Meyer, and
  Kollmeier]{schadler2012spectro}
Marc~Ren{\'e} Sch{\"a}dler, Bernd~T. Meyer, and Birger Kollmeier.
\newblock Spectro-temporal modulation subspace-spanning filter bank features
  for robust automatic speech recognition.
\newblock \emph{The Journal of the Acoustical Society of America}, 131\penalty0
  (5):\penalty0 4134--4151, 2012.

\bibitem[Schlag et~al.(2021)Schlag, Irie, and Schmidhuber]{schlag2021linear}
Imanol Schlag, Kazuki Irie, and J{\"u}rgen Schmidhuber.
\newblock Linear transformers are secretly fast weight programmers.
\newblock In \emph{Proc. ICML}, 2021.

\bibitem[Shi et~al.(2021)Shi, Wang, Wu, Yeh, Chan, Zhang, Le, and
  Seltzer]{shi2021emformer}
Yangyang Shi, Yongqiang Wang, Chunyang Wu, Ching-Feng Yeh, Julian Chan, Frank
  Zhang, Duc Le, and Mike Seltzer.
\newblock Emformer: Efficient memory transformer based acoustic model for low
  latency streaming speech recognition.
\newblock In \emph{Proc. ICASSP}, 2021.

\bibitem[Su et~al.(2024)Su, Ahmed, Lu, Pan, Bo, and Liu]{su2024roformer}
Jianlin Su, Murtadha Ahmed, Yu~Lu, Shengfeng Pan, Wen Bo, and Yunfeng Liu.
\newblock {RoFormer}: Enhanced transformer with rotary position embedding.
\newblock \emph{Neurocomputing}, 568, 2024.

\bibitem[Tillet et~al.(2019)Tillet, Kung, and Cox]{tillet2019triton}
Philippe Tillet, Hsiang-Tsung Kung, and David Cox.
\newblock Triton: an intermediate language and compiler for tiled neural
  network computations.
\newblock In \emph{Proc. MAPL}, 2019.

\bibitem[Watanabe et~al.(2018)Watanabe, Hori, Karita, Hayashi, Nishitoba, Unno,
  Soplin, Heymann, Wiesner, Chen, Renduchintala, and
  Ochiai]{watanabe2018espnet}
Shinji Watanabe, Takaaki Hori, Shigeki Karita, Tomoki Hayashi, Jiro Nishitoba,
  Yuya Unno, Nelson Enrique~Yalta Soplin, Jahn Heymann, Matthew Wiesner, Nanxin
  Chen, Adithya Renduchintala, and Tsubasa Ochiai.
\newblock {ESPnet}: End-to-end speech processing toolkit.
\newblock In \emph{Proc. Interspeech}, 2018.

\bibitem[Wijmans et~al.(2025)Wijmans, Huval, Hertzberg, Koltun, and
  Kr{\"a}henb{\"u}hl]{wijmans2025cce}
Erik Wijmans, Brody Huval, Alexander Hertzberg, Vladlen Koltun, and Philipp
  Kr{\"a}henb{\"u}hl.
\newblock Cut your losses in large-vocabulary language models.
\newblock In \emph{Proc. ICLR}, 2025.

\bibitem[Yang and Zhang(2024)]{fla2024}
Songlin Yang and Yu~Zhang.
\newblock {FLA}: A {Triton}-based library for hardware-efficient
  implementations of linear attention mechanism.
\newblock \url{https://github.com/fla-org/flash-linear-attention}, 2024.

\bibitem[Yang et~al.(2024)Yang, Wang, Zhang, Shen, and Kim]{yang2024deltanet}
Songlin Yang, Bailin Wang, Yu~Zhang, Yikang Shen, and Yoon Kim.
\newblock Parallelizing linear transformers with the delta rule over sequence
  length.
\newblock In \emph{Proc. NeurIPS}, 2024.

\bibitem[Yang et~al.(2025)Yang, Kautz, and Hatamizadeh]{yang2025gdn}
Songlin Yang, Jan Kautz, and Ali Hatamizadeh.
\newblock Gated delta networks: Improving {Mamba2} with delta rule.
\newblock In \emph{Proc. ICLR}, 2025.

\bibitem[Yao et~al.(2024)Yao, Guo, Yang, Kang, Kuang, Yang, Jin, Lin, and
  Povey]{yao2024zipformer}
Zengwei Yao, Liyong Guo, Xiaoyu Yang, Wei Kang, Fangjun Kuang, Yifan Yang,
  Zengrui Jin, Long Lin, and Daniel Povey.
\newblock Zipformer: A faster and better encoder for automatic speech
  recognition.
\newblock In \emph{Proc. ICLR}, 2024.

\bibitem[Yu et~al.(2021)Yu, Han, Gulati, Chiu, Li, Sainath, Wu, and
  Pang]{yu2021dualmode}
Jiahui Yu, Wei Han, Anmol Gulati, Chung-Cheng Chiu, Bo~Li, Tara~N. Sainath,
  Yonghui Wu, and Ruoming Pang.
\newblock Dual-mode {ASR}: Unify and improve streaming {ASR} with full-context
  modeling.
\newblock In \emph{Proc. ICLR}, 2021.

\bibitem[Zeghidour et~al.(2021)Zeghidour, Teboul, de~Chaumont~Quitry, and
  Tagliasacchi]{zeghidour2021leaf}
Neil Zeghidour, Olivier Teboul, F{\'e}lix de~Chaumont~Quitry, and Marco
  Tagliasacchi.
\newblock {LEAF}: A learnable frontend for audio classification.
\newblock In \emph{Proc. ICLR}, 2021.

\bibitem[Zeyer et~al.(2021)Zeyer, Schl{\"u}ter, and Ney]{zeyer2021peaky}
Albert Zeyer, Ralf Schl{\"u}ter, and Hermann Ney.
\newblock Why does {CTC} result in peaky behavior?
\newblock \emph{arXiv preprint arXiv:2105.14849}, 2021.

\bibitem[Zhang et~al.(2020{\natexlab{a}})Zhang, Wu, Yao, Wang, Yu, Yang, Guo,
  Hu, Xie, and Lei]{zhang2020unified}
Binbin Zhang, Di~Wu, Zhuoyuan Yao, Xiong Wang, Fan Yu, Chao Yang, Liyong Guo,
  Yaguang Hu, Lei Xie, and Xin Lei.
\newblock Unified streaming and non-streaming two-pass end-to-end model for
  speech recognition.
\newblock \emph{arXiv preprint arXiv:2012.05481}, 2020{\natexlab{a}}.

\bibitem[Zhang et~al.(2020{\natexlab{b}})Zhang, Lu, Sak, Tripathi, McDermott,
  Koo, and Kumar]{zhang2020transformer}
Qian Zhang, Han Lu, Hasim Sak, Anshuman Tripathi, Erik McDermott, Stephen Koo,
  and Shankar Kumar.
\newblock Transformer {Transducer}: A streamable speech recognition model with
  transformer encoders and {RNN-T} loss.
\newblock In \emph{Proc. ICASSP}, 2020{\natexlab{b}}.

\end{thebibliography}

\appendix
\section{Hyperparameters}
\begin{table}[h]
\centering\small
\resizebox{\ifdim\width>\linewidth\linewidth\else\width\fi}{!}{%
\begin{tabular}{lll}
\toprule
 & study model & production model \\
\midrule
layers, $d$, heads & 12, 256, 4 & 18, 1024, 16 \\
global pattern & GGGS $\times$ 3 & GGGS $\times$ 4 + GG \\
GDN-2 heads, head size, decay rank & 4, 64, 64 & 16, 64, 64 \\
GDN-2 output gate, RMSNorm $\epsilon$ & sigmoid, $10^{-6}$ & sigmoid, $10^{-6}$ \\
attention window & 128 frames & 128 frames, chunked lookahead $C \in \{1,2,7,14\}$ \\
cgMLP width, Gabor kernel & 1024, 31 & 4096, 31 \\
Gabor atoms & 2 Gaussians + 7 $\times$ (cos, sin) & same \\
front end & 128 mel, Conv2d $\times$8, 80\,ms & 128 mel, Conv2d $\times$8, 80\,ms \\
output & CTC, 5{,}000 BPE & RNN-T, 13{,}100 pieces + blank \\
prediction network & --- & 4 layers (G, A, G, A), $d{=}640$, attention window 128 tokens \\
optimizer & Muon + AdamW, lr $10^{-3}$, warmup 500 & Muon + AdamW, bf16 \\
\bottomrule
\end{tabular}}
\end{table}

\section{State accounting}
Production encoder, fp32: GDN-2 $14 \times 16 \times 64 \times 64 \times 4\,\mathrm{B} = 3.7$\,MB; attention windows
$4 \times 128 \times 2 \times 1024 \times 4\,\mathrm{B} = 4.2$\,MB; Gabor convolution caches
$18 \times (2048 + 2048) \times 30 \times 4\,\mathrm{B} = 8.8$\,MB; GDN-2 short-convolution caches 0.5\,MB.
Prediction network: GDN-2 $2 \times 10 \times 64 \times 64 \times 4\,\mathrm{B} = 0.3$\,MB and token-attention ring buffers
$2 \times 128 \times 2 \times 640 \times 4\,\mathrm{B} = 1.3$\,MB. A same-size encoder with unbounded causal attention in
all 18 layers caches $18 \times 2 \times 1024 \times 4\,\mathrm{B}$ per frame, 110\,MB per minute at 80\,ms frames.

\end{document}